\documentclass{article}

\PassOptionsToPackage{numbers, compress}{natbib}
    \usepackage[preprint]{neurips_2024}

\usepackage[utf8]{inputenc} % allow utf-8 input
\usepackage[T1]{fontenc}    % use 8-bit T1 fonts
\usepackage{hyperref}       % hyperlinks
\usepackage{url}            % simple URL typesetting
\usepackage{booktabs}       % professional-quality tables
\usepackage{amsfonts}       % blackboard math symbols
\usepackage{nicefrac}       % compact symbols for 1/2, etc.
\usepackage{microtype}      % microtypography
\usepackage{xcolor}         % colors
\usepackage{array}
\usepackage{booktabs}
\usepackage{graphicx}
\usepackage{amssymb}
\usepackage{pifont}

\title{EMGBench: Benchmarking Out-of-Distribution Generalization and Adaptation for Electromyography}

\author{%
  Jehan Yang\thanks{These authors contributed equally to this work.} \\
  Carnegie Mellon University\\
  Pittsburgh, PA 15213 \\
  \texttt{jehan@cmu.edu} \\
  \And
  Maxwell Soh$^*$ \\
  Carnegie Mellon University \\
  Pittsburgh, PA 15213 \\
  \texttt{msoh@andrew.cmu.edu} \\
  \And
  Vivianna Lieu \\
  Carnegie Mellon University \\
  Pittsburgh, PA 15213 \\
  \texttt{vlieu@andrew.cmu.edu} \\
  \And
  Douglas J Weber \\
  Carnegie Mellon University \\
  Pittsburgh, PA 15213 \\
  \texttt{dweber2@andrew.cmu.edu} \\
  \And 
  Zackory Erickson \\
  Carnegie Mellon University \\
  Pittsburgh, PA 15213 \\
  \texttt{zerickso@andrew.cmu.edu} \\
}

\newcommand{\cmark}{\ding{51}}%
\newcommand{\xmark}{\ding{55}}%

\newcommand{\greencheck}{{\color{green}\cmark}}

\newcommand{\redx}{{\textcolor{red}{\xmark}}}

\begin{document}

\maketitle

\begin{abstract}
  This paper introduces the first generalization and adaptation benchmark using machine learning for evaluating out-of-distribution performance of electromyography (EMG) classification algorithms. The ability of an EMG classifier to handle inputs drawn from a different distribution than the training distribution is critical for real-world deployment as a control interface. By predicting the user’s intended gesture using EMG signals, we can create a wearable solution to control assistive technologies, such as computers, prosthetics, and mobile manipulator robots. This new out-of-distribution benchmark consists of two major tasks that have utility for building robust and adaptable control interfaces: 1) intersubject classification, and 2) adaptation using train-test splits for time-series. This benchmark spans nine datasets, the largest collection of EMG datasets in a benchmark. Among these, a new dataset is introduced, featuring a novel, easy-to-wear high-density EMG wearable for data collection. The lack of open-source benchmarks has made comparing accuracy results between papers challenging for the EMG research community. This new benchmark provides researchers with a valuable resource for analyzing practical measures of out-of-distribution performance for EMG datasets. Our code and data from our new dataset can be found at \url{emgbench.github.io}.
\end{abstract}

\section{Introduction}

Electromyography (EMG) sensors detect muscle and motor neuron activity from the body, allowing for wearable gesture-based control of robots or devices. Particularly, EMG sensors can be used to sense intended hand or other body movements from people who are unable to move parts of their body due to injury or neurodegenerative disease~\cite{yang2023high, atzori2014electromyography, ting2021sensing, souza2024direct, bonilla2023progressive}. For people with upper or lower limb amputations, EMG-based prosthetic arms or legs can be controlled using the residual muscles from the remaining limb after amputation~\cite{atzori2014electromyography}. Additionally, for people with paralysis from stroke or spinal cord injury (SCI), EMG sensors can detect motor intent based on residual muscle fiber activity~\cite{souza2024direct, ting2021sensing, bonilla2023progressive}.

Several EMG datasets have been made publicly available, although many of them differ in regards to the hardware used and the placements of the sensors~\cite{atzori2014electromyography, du2017surface, cote2019deep, ozdemir2022dataset, krilova2019emg, jiang2021open}. Due to these common differences between EMG control interfaces, it is important to evaluate multiple EMG datasets to assess the classification accuracy and hardware-agnostic nature of machine learning-based techniques~\cite{sultana2023systematic, gohel2020review}. However, there is not yet a standardized benchmark for evaluating machine learning algorithm-based classification for EMG. This gap significantly impacts the standardization of results achieved from learning-based classification of EMG datasets, making results from machine learning papers that test on EMG difficult to compare. Furthermore, the absence of benchmarks designed to assess practical generalization and adaptation tasks, particularly those involving inter-subject performance evaluation, represents a notable gap in the current research landscape. In this work, we define and benchmark \textit{generalization} as the ability of a model to classify the gestures of a subject without using any of their data for training, and \textit{adaptation} as the ability of a model to personalize by fine-tuning using initial data from a subject after being pretrained on data from other subjects.

Improving performance on out-of-distribution subject generalization and adaptation-based tasks could significantly streamline the setup process for EMG interfaces, making them more accessible and easier to use for new users. For example, by benchmarking generalization tasks such as intersubject classification~\cite{wu2023Transfer, ctrl2024generic}, we can evaluate the performance of algorithms trained on large datasets to generalize to new subjects, enabling new users to control an interface without requiring additional data collection. In addition, by evaluating adaptation by fine-tuning using an initial subset of data from a new user~\cite{zhang2022self, ctrl2024generic}, we can determine the minimum amount of labeled data needed from a new user to personalize a model and achieve high gesture recognition accuracy~\cite{shi2024emg}. For the aim of creating a benchmark for out-of-distribution subject generalization and adaptation for EMG, while including some of the most popular EMG datasets, we have also curated a number of datasets that have been demonstrated to have strong performance for out-of-distribution generalization and adaptation~\cite{li2023deep, lu2022domain, cote2019deep}. In all, we present the following contributions:

\begin{figure}
\label{fig:datasets}
  \centering
  \includegraphics[width=\textwidth]{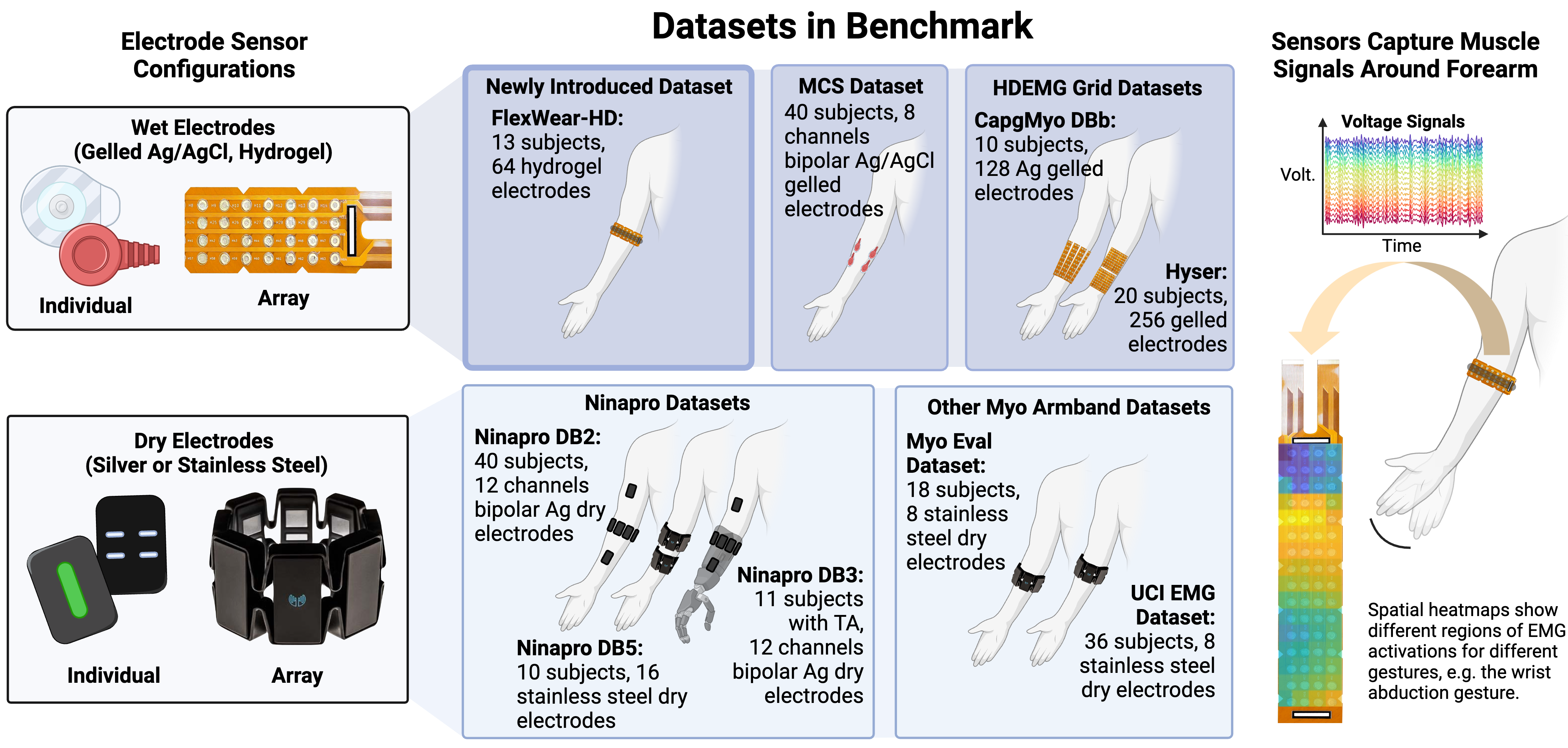}
  \caption{\textbf{Electrode configurations for various datasets and generalization task. } (Left) We show the main categories of electrode configurations used: dry electrodes and wet electrodes. These categories can be further separated into individually placed electrodes and electrode arrays.
  (Middle) The electrode configurations and placements used are shown for each dataset. We include nine total EMG datasets. Ninapro DB3 includes subjects with transradial amputations (TA). 
  (Right) We show that voltage signals from the arm are detected using EMG sensors and illustrate that gesture-specific patterns of muscle activity can occur.}
  \vspace{-2em}
\end{figure}

\textbf{The first open-source EMG benchmark} \hspace{0.5em} This benchmark presents a codebase for standardized evaluation of machine learning models on 9 curated EMG datasets for out-of-domain generalization and adaptation tasks. The codebase is available at \url{https://github.com/jehanyang/emgbench}. 

\textbf{New dataset with wearable EMG sensor} \hspace{0.5em} We present data using an easy-to-wear, reusable, high-density EMG sensor. Results evaluating the generalizability between subjects in this dataset show high classification accuracy. 

\textbf{Benchmarking results across tasks} \hspace{0.5em} Benchmarking results for a range of machine learning models and data preprocessing techniques across multiple generalization and adaptation tasks. 

\section{Background and Related Work}

\subsection{EMG signals}  

Although non-invasive EMG sensors are placed on the skin, these sensors can readily record significant voltage signals caused by the changes in voltage that occur following motor neuron action potentials. Muscles can amplify the biological voltage changes initiated by motor neurons~\cite{purves2001neuroscience}. This is because of the high number of moving ions during muscle fiber action potentials compared to neuronal action potentials, and the high number of muscle fibers activated per motor neuron, ranging from around 100 to as many as 1000~\cite{purves2001neuroscience}. These changes in voltage signals can be detected by skin electrodes, with both dry and wet electrodes used in common non-invasive EMG devices, as shown in Figure \ref{fig:datasets}. More details about EMG signals are in Appendix \ref{appendix:emg}. 

\subsection{EMG as a control interface}
\label{section:emg-as-a-control-interface}

EMG sensors have demonstrated the capability to detect signals from the muscles in an amputated forearm~\cite{maduri2019upper}, enabling high-dimensional control of prosthetic arms by leveraging residual muscle activity~\cite{lukyanenko2021stable}. Furthermore, as an alternative to motion-based interfaces~\cite{padmanabha2024independence, padmanabha2023hat}, in individuals with clinically motor complete cervical spinal cord injuries resulting in hand paralysis, machine-learning algorithms have shown promise in predicting voluntary hand gesture intentions at the individual finger level, given EMG signals from seven subjects with paralysis~\cite{souza2024direct}.

% For individuals without motor impairments, EMG-based sensors present a compelling alternative to camera-based systems for gesture control of robots and other devices~\cite{corradini2000camera, shaw2024learning}. This approach circumvents challenges associated with camera occlusion~\cite{han2018robust}, and can offer the advantage of detecting muscular force-based activation as additional inputs into an interface~\cite{corbett2011comparison}.
% Even for people without motor impairments, EMG-based sensors can be used as an alternative to camera-based~\cite{corradini2000camera, shaw2024learning} gesture control of robots or devices. This alternative does not face challenges for gesture-classification or pose estimation involving camera occlusion~\cite{han2018robust}, and can additionally detect muscular force-based activation as additional inputs into an interface~\cite{corbett2011comparison}. 

EMG often faces non-stationary signals that historically have made generalization difficult for learning-based methods. However, these instances present a unique opportunity for domain adaptation methods. EMG signals experience several phenomena that cause \textbf{concept shifts}, altering the conditional probability of labels given the inputs from the training set to the test set~\cite{campbell2024context}. The mechanisms causing this phenomenon include variations in muscle locations~\cite{dellon1987musculoaponeurotic}, arm sizes~\cite{nourbakhsh2004relationship}, skin impedances~\cite{rask2019genome}, and electrode placements~\cite{mesin2009surface}. By accounting for some of these mechanisms through fine-tuning or other methods, adaptation can potentially maintain or improve classification performance between subjects.

\subsection{Classification over EMG datasets}

Extensive research has focused on training machine learning models for EMG-based gesture classification, utilizing both publicly available datasets~\cite{lu2022domain, islam2024surface, wei2021hierarchical, hye2023artificial} and novel datasets collected by the researchers~\cite{cote2019deep, yang2023high, ozdemir2022hand, li2023deep, wang2023pruning, zhang2023multi, xu2023cross, alguner2023window, ctrl2024generic}. While many studies report results based on randomized train-test split accuracy~\cite{hye2023artificial, alguner2023window, sri2021classification} and k-fold cross-validation (KFCV)~\cite{zhang2022research, ozdemir2022hand, fatimah2021hand, he2020biometric, kim2019development}, where data from the training, validation, and test sets may be randomly sampled from the same subjects, such approaches may not accurately reflect real-world scenarios. In practice, it is often desirable for the validation and test sets to comprise data collected either temporally after the training data from the same subject (termed train-test splits for time series, or TSTS), or from a subject entirely excluded from the training set, as evaluated using leave-one-subject-out cross-validation (LOSO-CV). These data splitting strategies provide more robust assessments of model performance by introducing out-of-distribution generalization challenges. We present a categorization on how several other EMG classification papers split their data in Appendix Table \ref{tab:data-splits}.

In the case of TSTS, we test with data collected after the training set, which introduces potential distribution shifts due to factors such as variations in gesture execution, fatigue~\cite{liu2021muscle, chua2024analysis}, perspiration~\cite{abdoli2012effect}, electrode displacement on the skin~\cite{de1996origin}, drying or changes in ionic concentrations of hydrogel or electrolyte gels~\cite{sousa2023long}, and variations in electrode adherence on the skin~\cite{chi2010dry}. Similarly, LOSO-CV introduces variability stemming from inter-individual differences in body size, muscle morphology~\cite{dellon1987musculoaponeurotic}, and differences in skin impedance and adipose tissue distribution~\cite{rask2019genome}. Studies employing randomized or mixed data splits, where evaluation data may precede training data, risk reporting artificially inflated accuracies that fail to reflect true generalization capabilities in practical EMG classifier deployments. To facilitate benchmarking, a standardized approach to EMG sample window extraction from raw data can ensure consistent data preprocessing across studies~\cite{kulwa2022analyzing, sultana2023systematic}. This methodology enables more reliable comparisons of classification performance. A detailed analysis of prior work on EMG generalization and adaptation is provided in Appendix~\ref{appendix:generalization-prior-work}.

\section{Datasets}

Several datasets have been released for classification of gestures using EMG. Between them, there are a large variety of experimental conditions and data collection protocols: some of these variations include the amount of time and repetitions used in cues for the participant to perform gestures~\cite{atzori2014electromyography, pizzolato2017comparison, ozdemir2022hand, wei2021hierarchical}, different numbers of participants, and different devices used to collect data~\cite{atzori2014electromyography, pizzolato2017comparison, ozdemir2022dataset}. Figure~\ref{fig:datasets} illustrates the sensor configurations and the number of subjects for each dataset, while Table \ref{tab:gesture-times} summarizes the amount of time the participant is cued to perform gestures. Further detail about each dataset is included in Appendix \ref{appendix:additional-details-on-datasets} as well as in Table \ref{tab:emg_datasets}. 

\textbf{Ninapro} \hspace{0.5em} One of the most popular EMG datasets, Ninapro includes over 180 data acquisition sessions and is subdivided into 10 sub-datasets~\cite{atzori2014electromyography, pizzolato2017comparison}. We include some of the most popular sub-datasets used for benchmarking machine learning algorithms: Ninapro DB2, DB3, and DB5~\cite{pizzolato2017comparison, sultana2023systematic}.  The Ninapro DB5 dataset uses two sets of a low-cost wireless wearable EMG device called the Myo Armband~\cite{pizzolato2017comparison} worn on the same arm. Ninapro DB2 and DB3 use individually placed dry electrodes. All users in Ninapro DB3 have transradial amputations~\cite{atzori2014electromyography}.

\textbf{CapgMyo} \hspace{0.5em} Introduced in \citet{geng2016gesture} and further described in \citet{du2017surface}, the CapgMyo dataset includes 3 sub-datasets: DB-a, DB-b, and DB-c. We include DB-b in our benchmark dataset, which has multi-session data. In total, CapgMyo DB-b includes 10 subjects performing 8 gestures and data recorded by 128 high-density electrodes, separated into 8 acquisition modules with 16 electrodes per module. 

\textbf{Myo Dataset} \hspace{0.5em} Described in \citet{cote2019deep}, the Myo dataset uses the wireless wearable EMG device called the Myo Armband. Like in \cite{lin2020normalisation}, we include the evaluation dataset, which has 18 subjects. In this dataset, 7 different gestures are recorded and the methodology for placing the Myo Armband on the arm is specified in detail. 

\textbf{EMG Data for Gestures Dataset} \hspace{0.5em} The EMG Data for Gestures (UCI EMG) dataset~\cite{krilova2019emg}, hosted on the UC Irvine Machine Learning Repository, has been used to benchmark leave-subject-out tests~\citet{lu2022domain}. The device used is the Myo Armband. This dataset includes 36 subjects and includes 2 sessions, with only 1 gesture performance for each of 6 or 7 gestures performed per session. Although the timespan between the two sessions is not specified, both sessions for each participant are collected on the same day, as indicated by date-based timestamps in the filenames.

\textbf{Multi-channel sEMG Dataset} \hspace{0.5em} The multi-channel sEMG (MCS) dataset presented in \citet{ozdemir2022dataset} uses 4 bipolar Ag/AgCl electrode channels individually placed on 40 subjects' arms, placed by approximate locations of the specific muscles. Electrolyte gel is placed on the arm under the Ag/AgCl electrodes. The muscles measured from are the extensor carpi radialis, flexor carpi radialis, extensor carpi ulnaris, and flexor carpi ulnaris. 

\textbf{Hyser Dataset} \hspace{0.5em} The Hyser dataset, with details specified in \citet{jiang2021open}, includes data from 20 subjects and 34 gestures while recording from 256 gelled electrodes separated into 4 grids of high density electrodes, with two grids on the flexor side and two grids on the extensor side. Although there does not seem to be details on the exact materials used in the device, the most common gelled electrode materials used are Ag/AgCl along with a Cl electrolyte gel. 

\textbf{FlexWear-HD Dataset} \hspace{0.5em} We present a 13-person EMG dataset using a reusable electrode array called the FlexWear-HD dataset. This array uses a flexible printed circuit board (FPCB) with 64 hydrogel electrodes placed onto gold-plated copper pads on the FPCB. Two sessions are presented per subject, with the time between each session being about 1 hour. The wearable array is also kept on between the two sessions, allowing for the evaluation of typical changes that occur over time on EMG signals without the effects that can occur from replacing the electrode array on the arm. 

\begin{figure}
\label{fig:training-pipeline}
  \centering
  \includegraphics[width=\textwidth]{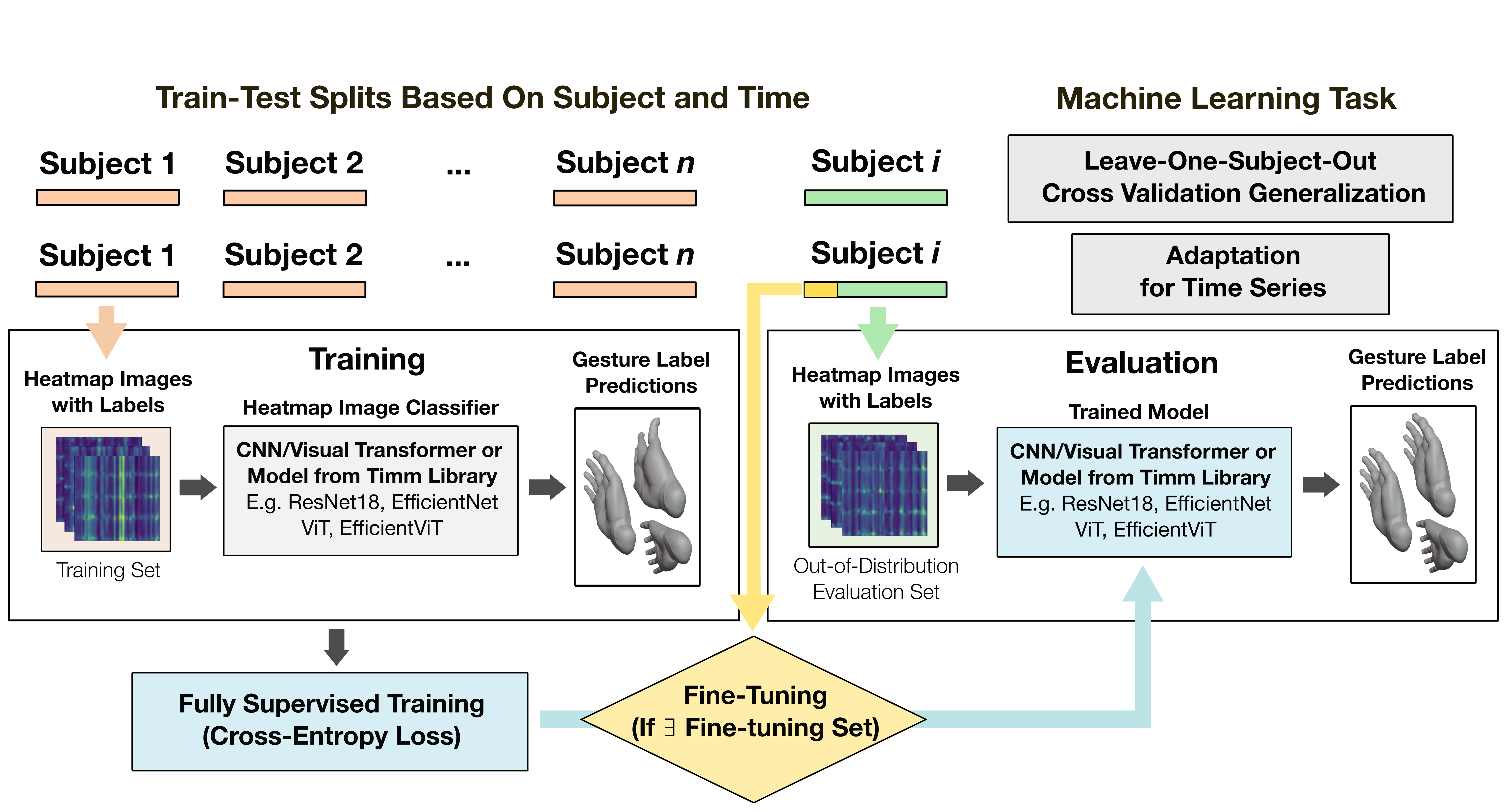}
  \caption{\textbf{Training pipeline for testing generalization and adaptation.} We show the training and evaluation pipeline for our benchmarking script to evaluate generalization across subjects and over time. For \textit{leave-one-subject-out cross validation} (LOSO-CV), the training set involves all subjects other than the test subject, with the left-out subject $i$ changing from $1$ to $n$ between training runs. In addition, we test for \textit{adaptation for time-series}, where data from the beginning of subject $i$ is used for fine-tuning after pretraining a model on data from other subjects. In our benchmark, both \textit{few-shot fine-tuning} and \textit{intersession fine-tuning} are used to evaluate adaptation for time-series.}
  \vspace{-1em}
\end{figure}

% Methods section
\section{Methods}

This section outlines our methods for developing and evaluating gesture classification for EMG. We describe preprocessing techniques to convert time-series EMG data into 2D activity maps, the gesture classification models tested, and the generalization tasks for cross-subject and cross-session performance. Additionally, we detail the classification metrics and hardware setup used in our experiments. We note that we always use a constant numbers of epochs for training the classification models. 

\subsection{Preprocessing Methods}
\label{sect:preprocessing-methods}
A variety of preprocessing methods have been proposed for model training on EMG data. Given the time-series nature of EMG data, feature extraction methods include root-mean-square (RMS), number of zero-crossings, and mean absolute value~\cite{phinyomark2012feature}. By converting raw time series data or other manual features over time to heatmaps, we are able to create spatiotemporal patterns from time-series data, which can be given as input into 2D CNNs. We convert time series data for each electrode into separate rows in the activity map. Another approach to convert time-series data into a 2D format involves time-frequency transforms, such as spectrograms and continuous wavelet transforms (CWT). In this benchmark, we evaluate preprocessing using 1) heatmaps from raw data, 2) heatmaps from RMS windows, 3) spectrograms, and 4) CWTs. We show examples of these preprocessing methods resulting in activity maps in Figure \ref{fig:preprocessing_methods}. Classification of EMG data using 2D CNNs after preprocessing has achieved high accuracy in prior studies~\cite{geng2016gesture, ozdemir2022hand}. We review ways that EMG has been processed into activity maps in Appendix \ref{appendix:preprocessing}.

\subsection{State-of-the-Art Image Classifier Algorithms}

Several image classifier models have been successfully applied to EMG data~\cite{ozdemir2022hand, geng2016gesture, dere2023novel, montazerin2023transformer}. Following the examples of prior EMG classification work by \citet{ozdemir2022hand} and \citet{dere2023novel}, we evaluate on the ImageNet-pretrained ResNet18, and Visual Transformer (ViT) models~\cite{he2016deep, wu2022tinyvit}. Additionally, we test the EfficientNet and EfficientViT models, which have shown strong performance on ImageNet classification while using fewer model parameters or achieving faster inference compared to other state-of-the-art visual models~\cite{tan2020efficientnet, cai2022efficientvit}. The number of parameters used for each model is 2 million for EfficientViT, 4 million for EfficientNet, 6 million for ViT, and 11 million for Resnet18. By utilizing the PyTorch Image Models library in our benchmarking code, we enable other researchers to easily benchmark different visual classifier models by simply modifying the configuration file, with support for both pretrained and untrained models.

\subsection{Different Generalization and Adaptation Tasks}

Generalization and adaptation across different setups is essential for robust EMG-based models for real-world deployment. We evaluate two tasks: 1) generalization on a left-out subject, and 2) adaptation using initial data from a subject. For this second task, we fine-tune a pretrained model using initial data from the evaluation subject or the first session from the evaluation subject, respectively.

In task 1, we use LOSO-CV to assess how well an EMG gesture classifier and interface may work for a new user whose data was not included in training. In task 2, we perform TSTS on the data data to assess few-shot learning through fine-tuning, determining the minimal data required from a new subject for robust classification. We also test intersession accuracy to evaluate how well an EMG interface works for a user in a new session without recalibration. Train-test splits for generalization and adaptation are illustrated in Figure \ref{fig:training-pipeline}. More details on the classification metrics and data splits are provided in Appendix \ref{appendix:classification-metrics}. 

\subsection{Hardware Used for Benchmarking}
\label{section:hardware}

In our benchmarking experiments, we use GPU nodes from the Pittsburgh Supercomputer Center, which have eight NVIDIA Tesla V100-32GB SXM2 GPUs each. Each node uses two Intel Xeon Gold 6248 ``Cascade Lake'' CPUs, and 512GB of DDR4-2933 RAM. In total, our tests took around 10,000 GPU hours. Much of the compute required comes from training individual models for each individual subject in order to evaluate average classification metrics across subjects. 

\section{Results}
\label{section:results}

In this section, we present the outcomes of our experiments, highlighting the generalization and adaptation performance of gesture classification models using EMG. We evaluate the effectiveness of various data preprocessing techniques, followed by benchmarking different machine learning architectures. We then vary the amount of data used for fine-tuning, and present results on generalizing and adapting to data that comes from multiple sessions. The results provide insights into the generalization and adaptation tasks across different datasets, offering insight into the robustness and applicability of gesture classifiers in real-world scenarios. 

% \subsection{Distance Matrix Showing Distribution Shift}

% Distance matrices can be used to analyze the similarity of data across gestures and sessions, highlighting any distribution shifts that occur. In Figure \ref{fig:distance-matrices}, we present two distance matrices for the FlexWear-HD dataset. The left matrix shows significant differences between the average signals of each gesture across the two sessions, particularly in the bottom left quadrant. The right matrix reveals variations over time between samples of the same gesture, with intrasession differences visible near the diagonal.

% \begin{figure}
% \label{fig:distance-matrices}
    
% \includegraphics[width=\columnwidth]{images/distance-matrices.png}
% \caption{\textbf{Comparison of gesture data similarity between the first and second sessions of data collection.} (Left) A Euclidean distance matrix displays distances between the mean RMS heatmaps for all gesture types, normalized by electrode, using data from all participants in the FlexWear-HD dataset. (Right) Another Euclidean distance matrix shows distances between each sample for the first user in the FlexWear-HD dataset, highlighting shifts that can occur for the same gesture over multiple trials.}
% \vspace{-1em}
% \end{figure}

\subsection{Data Representation Benchmarking}

Generalization and few-shot fine-tuning adaptation results with raw heatmap images, RMS heatmap images, spectrograms, and CWT preprocessing methods are presented in Table \ref{tab:preprocessing-vs-datasets}. Each value represents the mean gesture recognition performance averaged over all $N$ models and train-test splits from LOSO-CV, where $N$ is the number of subjects in a given dataset. Previous studies~\cite{ozdemir2022hand, nahid2020deep} have demonstrated strong gesture recognition performance with an ImageNet pre-trained ResNet model. Similarly, we use the ImageNet pre-trained ResNet-18 model for our experiments to benchmark different preprocessing methods. For all preprocessing methods and datasets, we pretrain a model using data from all training subjects, then fine-tune it using the first 20\% of data from the left-out evaluation subject, with data splits stratified by each gesture. A flow-chart showing this process is shown in Figure \ref{fig:training-pipeline}. Following this fine-tuning process, test accuracy performances for all four preprocessing methods are significantly higher than those for LOSO-CV, illustrating how fine-tuning using a small amount of data from a left-out subject can significant improve classification accuracy. 

\begin{figure}
\label{fig:preprocessing_methods}
  \centering
  \includegraphics[width=1\textwidth]{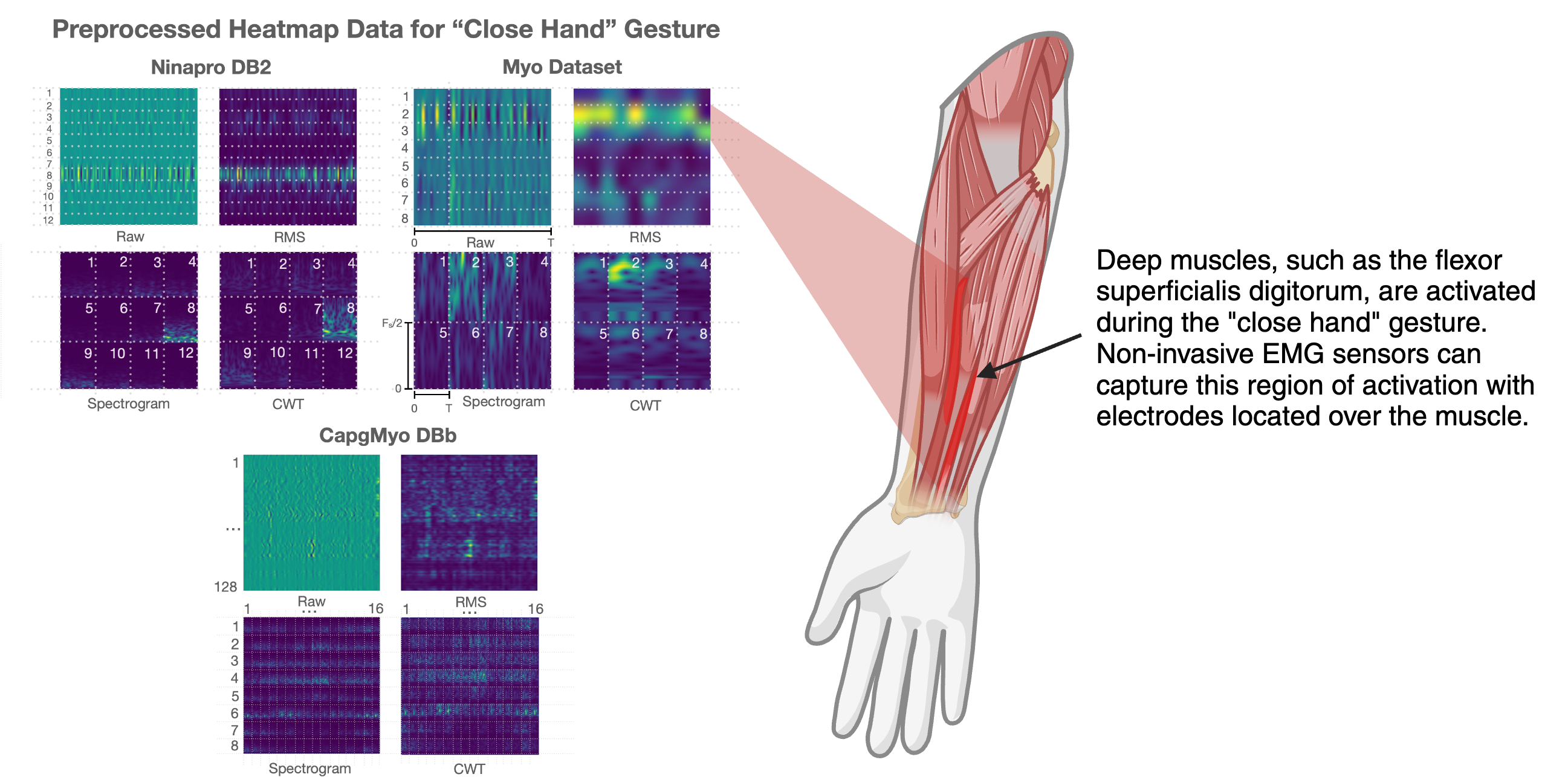}
  \caption{\textbf{Varying preprocessing methods for generating heatmaps.} Samples from different preprocessing methods are shown for the 128 electrode CapgMyo dataset, the 12 electrode NinaPro DB2, and the 8 electrode Myo Dataset. Values on the heatmaps correspond to the index of the electrode for the sub-image shown on the grid. All samples correspond to the closed-hand gesture. The closed-hand gesture primarily activates deep muscles of the flexor side of the forearm, such as the flexor digitorum superficialis.}
\end{figure}

Raw heatmaps, spectrograms, and CWTs all perform well for most datasets, with different datasets having different preprocessing methods that work the best out of the four. The largest difference in performance between the preprocessing methods occurs in the MCS and Hyser datasets, with time-frequency transforms working significantly better than raw heatmaps. Previous work has not explored how the Hyser dataset performs when comparing the classification of both raw heatmaps and time-frequency transforms, with \citet{li2023deep} only exploring raw heatmaps. Similarly, previous work~\cite{ozdemir2022hand} on the MCS dataset has only evaluated CWT and STFT, and not raw or RMS-based preprocessing. There is a lack of work showing how the phase from time-frequency transforms, such as the STFT and the Hilbert-Huang transform performs as a preprocessing method for EMG-based gesture classification. We present these results in Appendix \ref{appendix:phase-preprocessing}, but note that we do find these phase-based preprocessing method performances to be significantly lower than when preprocessing using the magnitude of STFT or CWT. The reason for the decrease in performance may be due to the lack of any magnitude-based information, which can often differentiate gestures based on significant EMG activity from specific electrodes, illustrated in Figure \ref{fig:preprocessing_methods}. For the experiments in Table \ref{tab:preprocessing-vs-datasets}, pretraining is conducted over 100 epochs and fine-tuning is conducted over 750 epochs.

\begin{table}[t]
    \centering
    {
    \tiny
    \begin{tabular}{lccccccccc}
        \toprule
            & \bf{Myo Dataset}  & \bf{UCI}          & \bf{NinaproDB5} & \bf{Capgmyo}  & \bf{NinaproDB2} & \bf{NinaproDB3} & \bf{MCS}           & \bf{Hyser}          & \bf{FlexWear-HD}   \\
        \midrule
        \multicolumn{10}{l}{\bf{Pretraining before few-shot fine-tuning, LOSO-CV}} \\
        \midrule
        Raw & \bf{74.5/94.8}    &        76.4/94.7  & \bf{41.3/79.4} &     42.4/81.0  &  18.4/62.0      &      11.2/52.0  &         67.7/93.2  &          44.4/82.1  &     \bf{77.7/97.4} \\
        RMS &         68.7/92.5 &        75.4/94.5  &     39.2/76.1  &     41.8/81.4  &  17.2/60.1      &      11.2/52.8  &         58.2/88.0  &          48.5/83.5  &         75.6/97.2  \\
        STFT&         72.8/94.3 &    \bf{78.2/95.6} &     41.3/78.1  &     41.9/80.4  &  19.2/62.6      &  \bf{12.0/52.8} &         74.4/95.6  &          54.4/88.9  &         75.3/96.8  \\
        CWT &         69.7/92.6 &        75.9/94.1  &     38.2/77.2  & \bf{42.9/81.5} &  \bf{20.1/63.2} &      10.9/51.6  &     \bf{77.9/96.7} &      \bf{58.0/90.4} &         75.1/97.2  \\
        \midrule
        \multicolumn{10}{l}{\bf{Finetuning with first 20\% of data from left-out subject}} \\
        \midrule
        Raw &   \bf{95.1/99.5} &        91.6/99.2   & \bf{68.3/94.1} & \bf{92.8/99.5} &  52.2/86.0      &       44.2/80.0 &          89.5/98.5 &          79.1/96.0  &         95.2/99.6  \\
        RMS &       94.1/99.3  &        91.2/99.2   &     67.7/93.4  &     89.4/99.0  &  49.4/84.5      &      40.6/77.5  &          78.0/95.6 &          84.8/89.1  &         95.7/99.7  \\
        STFT&       95.0/99.5  &     \bf{91.7/99.1} &     66.6/93.1  &     92.2/99.5  &  \bf{53.1/86.5} &  \bf{44.7/80.2} &          90.9/99.0 &          89.4/98.6  &    \bf{95.8/99.7}  \\
        CWT &       92.5/99.0  &    \bf{91.7/98.8}  &     62.1/91.0  &     89.7/99.2  &  51.9/85.9      &      43.4/79.6  &     \bf{92.1/99.1} &      \bf{90.3/98.8} &         95.6/99.8  \\
        \bottomrule
    \end{tabular}

    }
    \vspace{1em}
    \caption{\label{tab:preprocessing-vs-datasets}\textbf{Benchmark of EMG Preprocessing Techniques.} LOSO-CV average test accuracy (Acc) / area under the receiver operating characteristic (AUROC) for all the datasets and for four common EMG-to-image preprocessing methods: original temporal EMG signals (Raw), root-mean-square (RMS), short-time fourier transform (STFT), and continuous wavelet transform (CWT), as mentioned in Section \ref{sect:preprocessing-methods}. RMS consists of taking consecutive windows of length 16-20 timesteps and applying the RMS transform.}
    \vspace{-0.5em}
\end{table}

\subsection{Machine Learning Architecture Benchmarking}

Evaluating a variety of machine learning architectures is important for identifying the most effective models for EMG gesture classification, given heatmaps, spectrograms, and CWTs. We evaluate both convolutional neural networks and visual transformers due to their unique mechanisms: CNNs leverage convolutional and pooling layers to learn locally shift-invariant features~\cite{he2016deep}, while transformers use self-attention modules to model complex spatial relationships within an entire image~\cite{wu2022tinyvit}. We benchmark several gesture classification architectures that have demonstrated strong performance on EMG data~\cite{ozdemir2022hand, dere2023novel, nahid2020deep}. These architectures include Convolutional Neural Networks (CNNs)~\cite{ozdemir2022dataset, nahid2020deep, geng2016gesture} and vision transformers (ViTs)~\cite{dere2023novel} that are pretrained on the ImageNet dataset~\cite{deng2009imagenet} with all model weights unfrozen for training. We run all pretraining over 50 epochs, and fine-tuning over 375 epochs. For each dataset, we choose the preprocessing method that performs the best in leave-one-subject-out cross validation from Table \ref{tab:preprocessing-vs-datasets} ( shown in bold in the first four rows) and evaluate leave-one-subject-out cross-validation and few-shot transfer learning for each dataset for four image classifier models: 1) ResNet18, 2) EfficientNet, 3) ViT, and 4) EfficientViT. For all these models, we train from an ImageNet-1k pretrained model~\cite{deng2009imagenet}.

ResNet18 performs the best for Ninapro DB5, Ninapro DB2, and MCS. EfficientNet shows the best performance for CapgMyo, UCI, Ninapro DB3 and FlexWear-HD, while EfficientViT performs best for the Myo Dataset, and Hyser. Although, to our knowledge, the EfficientNet and EfficientViT model architectures have not previously been used to classify EMG signals, they have demonstrated state-of-the-art performance and inference speed for computer vision tasks in prior studies~\cite{cai2022efficientvit, tan2020efficientnet}.

As shown in both Tables \ref{tab:preprocessing-vs-datasets} and \ref{tab:datasets-vs-models}, datasets that use array-based wearable electrode devices such as the Myo Dataset, UCI EMG dataset, and the FlexWear-HD tend to perform better in leave-one-subject-out cross validation (LOSO-CV) when compared to datasets that require individually placed electrodes or electrode grids in uniform patterns around the arm, such as in CapgMyo, or Hyser. The exception to this seems to be the Ninapro DB5 and MCS datasets. For Ninapro DB5, although data is recorded with two Myo Armband wearables, it has low classification performance for LOSO-CV. We note that this may be due to having noisy labels~\cite{chang2020assessment}, which may cause artificially low generalization performance. For the MCS dataset, although this dataset has the fewest electrodes, it shows relatively high LOSO-CV performance. This may be because the MCS dataset uses individual electrodes that are placed by a researcher based on specific muscles, potentially decreasing the concept shift across subjects based on electrode placements~\cite{ozdemir2022dataset}. 

\begin{table}[h]
    \centering
    {
    \tiny
    \begin{tabular}{lccccccccc}
        \toprule
        & \bf{Myo Dataset} & \bf{UCI}       & \bf{NinaproDB5} &   \bf{Capgmyo} & \bf{NinaproDB2} & \bf{NinaproDB3} & \bf{MCS}       & \bf{Hyser}     & \bf{FlexWear-HD} \\
        \midrule
        \multicolumn{10}{l}{\textbf{Before few-shot finetuning through pretraining using leave-one-subject-out cross validation}} \\
        \midrule
        RN18       &       73.7/94.9  &     76.6/95.0  & \bf{42.1/78.9}  &     41.4/82.9  & \bf{19.9/63.0}  &      11.1/51.6  & \bf{77.8/96.6} &     56.6/89.1  &     77.0/97.4   \\
        EN   &       73.5/94.2  &  \bf{78.6/95.3}&     39.9/78.6   & \bf{46.6/84.2} & 19.5/63.0       &  \bf{11.6/52.3} &     76.9/96.4  &     58.6/90.7  & \bf{83.9/98.5}  \\
        ViT            &       70.8/93.2  &     77.3/95.1  &     38.8/77.7   &     39.8/78.9  & 18.0/61.1       &      11.1/51.3  &     76.3/96.0  &     56.3/88.8  &     70.2/96.3   \\
        EViT   &   \bf{73.8/93.9} &     77.2/94.4  &     40.9/78.7   &     45.0/82.8  & 18.8/62.0       &      11.1/51.3  &     76.2/96.2  & \bf{61.8/92.2} &     74.9/96.3   \\
        \midrule
        \multicolumn{10}{l}{\textbf{After few-shot fine-tuning by doing training using first 20\% of data from left-out subject}} \\
        \midrule
        RN18       &       94.6/99.5  &     91.4/99.0  &      68.5/93.6  & \bf{91.2/99.3} & 52.4/85.8       &  \bf{45.2/81.3} &     91.0/99.0  & \bf{87.4/98.4} &     95.9/99.8   \\
        EN   & \bf{94.9/99.6}   & \bf{91.7/99.1} &      67.7/93.3  &     89.7/98.6  & 53.8/87.1       &      45.1/81.0  &     91.0/99.0  &     85.4/97.8  & \bf{96.9/99.8}  \\
        ViT            &     94.9/99.4    &     90.1/98.1  &      68.2/94.2  &     85.1/98.5  & 51.0/86.1       &      41.2/78.5  &     91.4/98.9  &     70.6/93.3  &     93.6/99.5   \\
        EViT   &     94.9/99.4    & \bf{91.7/99.1} &  \bf{69.2/94.1} &     87.1/98.3  & \bf{54.0/86.9}  &      44.8/81.4  & \bf{92.0/98.8} &     83.2/96.6  &     95.3/99.6   \\
        \bottomrule
    \end{tabular}
    }
    \vspace{1em}
    \caption{\textbf{Benchmarking of Machine Learning Architectures.} Performance of gesture recognition models on each dataset (Acc/AUROC). The models used are the ImageNet-pretrained ResNet18 (RN18), EfficientNet (EN), ViT, and EfficientViT (EViT).}
    \label{tab:datasets-vs-models}
\end{table}

\subsection{Varying Amount of Data Available for Training}

Using the best performing pre-processing method found in Table \ref{tab:preprocessing-vs-datasets} and the corresponding best classifier architecture found in Table \ref{tab:datasets-vs-models} for each dataset, we evaluate adaptation methods in Table \ref{tab:adaptation-tasks}. For each of the FT-$X\%$ rows, we evaluate fine-tuning using the first $X\%$ of data from the left-out subject after pretraining a model on data from all other subjects. We stratify by gesture to achieve a balanced fine-tuning set. By varying the amount of data used for adaptation, we can determine the data required for significant performance increases.

Our findings indicate that for many datasets, performance increases with more fine-tuning data from the participant. However, for datasets such as the Myo Dataset and FlexWear-HD, performance reaches 93.1\% for 7 gestures and 94.2\% for 10 gestures, respectively, with just 5\% of a subject's initial data. Specifically, for the Myo dataset and FlexWear-HD dataset, 5\% of data equates to approximately eight seconds and 30 seconds of data, respectively. This suggests that only a short data collection time from a new subject is needed for high classification accuracy by fine-tuning a model pretrained on a large dataset of training subjects. In fact, the average percent change in accuracy compared to the pretrained model after fine-tuning using 5\% of the leftout subject's data was 44\%. A large-scale study by \citet{ctrl2024generic} found a similar trend, where personalizing a model by fine-tuning improves performance on EMG tasks by 30\%. For Table \ref{tab:adaptation-tasks}, we run the same experiments using 3 seeds and report the mean and standard deviation of test accuracies, finding that the variance in performance metrics is small.

For the intersession tests shown in Table \ref{tab:intersession}, we pretrain using data from all available sessions for the training subjects and then fine-tune using all the data from the first session of the left-out subject. This is in contrast to the evaluations in Tables \ref{tab:preprocessing-vs-datasets}, \ref{tab:datasets-vs-models}, and \ref{tab:adaptation-tasks}, where only the first session is used for training, fine-tuning, and evaluation sets. Note that in Table \ref{tab:intersession}, only 4 datasets are represented because only these datasets have more than data from 2 sessions. As Capgmyo and Hyser have sessions that are on different days, Table \ref{tab:intersession} shows significantly decreased performance after fine-tuning than in Table \ref{tab:adaptation-tasks}, where fine-tuning data comes from the same day as the test data. Additionally, for datasets that include transition data, Table \ref{tab:transition-tasks} shows performance when using data windows that capture both isometric holds and transitions between gestures. As anticipated, most datasets exhibit lower accuracies compared to Table \ref{tab:adaptation-tasks}, as the transition data introduces greater variability in EMG signals due to dynamic movements.

All pretraining is conducted over 50 epochs, and fine-tuning over 375 epochs. 

\begin{table}[h]
    \centering
    {
    \scriptsize
    \begin{tabular}{lccccc}
        \toprule
        \textbf{Task}    & \bf{Myo Dataset}    & \bf{UCI EMG}        & \bf{Ninapro DB5}     & \bf{Capgmyo} & \bf{Ninapro DB2} \\
        \midrule
        FT-5\%           & 93.2/99.3 (0.1/0.1) & 82.1/96.7 (0.8/0.3) & 48.4/83.7 (0.6/0.3)  & 76.1/95.9 (2.2/0.1) & 37.1/76.4 (0.4/0.3) \\ 
        FT-20\%          & 95.0/99.4 (0.3/0.1) & 91.7/99.1 (0.5/0.1) & 69.7/94.1 (1.1/0.4)  & 89.2/98.7 (0.9/0.2) & 52.3/86.0 (0.4/0.2) \\
        FT-40\%          & 98.3/99.9 (0.1/0.0) & 92.5/99.3 (0.3/0.1) & 76.0/96.4 (0.6/0.2)  & 96.1/99.7 (0.3/0.1) & 55.9/88.8 (0.3/0.2) \\
        FT-60\%          & 98.9/100.0 (0.1/0.1) & 95.0/99.6 (0.3/0.1) & 82.0/97.8 (0.1/0.1) & 96.4/99.9 (0.4/0.2) & 56.4/89.1 (0.7/0.2) \\
        FT-80\%          & 99.6/100.0 (0.1/0.0) & 97.4/99.8 (0.9/0.1) & 78.1/96.9 (0.6/0.3) & 97.7/99.9 (0.6/0.1) & 56.0/90.1 (0.7/0.3) \\
        % IS FT            & --                 & 93.5/99.6      & --               & 62.1/91.4    & --               \\
        \cmidrule[0.3mm](lr){1-6}
    \end{tabular}
    \scriptsize
    \begin{tabular}{lcccc}
        % \toprule
        \textbf{Task}    & \bf{Ninapro DB3} & \bf{MCS}       & \bf{Hyser}       & \bf{FlexWear-HD} \\
        \midrule
        FT-5\%           & 32.7/72.0 (0.5/0.7) & 81.1/96.5 (0.9/0.3) & 66.5/91.1 (2.8/1.3) & 94.6/99.6 (0.7/0.1) \\ 
        FT-20\%          & 44.4/80.3 (0.3/0.2) & 91.4/99.0 (0.7/0.1) & 84.1/96.5 (1.3/0.2) & 96.8/98.8 (0.2/1.6) \\
        FT-40\%          & 49.6/83.2 (0.3/0.2) & 95.6/99.7 (0.1/0.0) & 91.2/98.3 (1.7/0.5) & 98.9/99.9 (0.1/0.1) \\
        FT-60\%          & 51.2/85.1 (0.8/0.4) & 96.6/99.8 (0.3/0.1) & 92.0/98.5 (0.7/0.5) & 98.8/99.8 (0.3/0.0) \\
        FT-80\%          & 52.7/87.3 (1.1/0.5) & 97.2/99.8 (0.2/0.1) & 96.7/99.0 (0.4/0.4) & 99.3/100.0 (0.2/0.0) \\
        % IS FT            & --                 & --             & 60.1/91.2        & 99.1/100.0       \\
        \bottomrule
    \end{tabular}
    }
    \vspace{1em}
    \caption{\textbf{Benchmarking Amount of Data for Fine-tuning.} Performance on datasets for adaptation tasks useful in EMG control interfaces (Mean Acc/Mean AUROC with standard devations in parentheses).  LOSO-CV stands for leave-one-subject-out cross validation. IS FT stands for intersession fine tuning. FT-X\% involves using the first X\% of data from the left-out subject for fine-tuning after pretraining with data from all others subjects.}
    \label{tab:adaptation-tasks}
\end{table}

\begin{table}[h]
    \centering
    \small
    \begin{tabular}{lccccccccc}
    \toprule
    \textbf{Task} & \textbf{UCI}  & \textbf{Capgmyo}  & \textbf{Hyser} & \textbf{FlexWear-HD} \\
    \midrule
    IS w/o FT     & 79.8/95.5     & 47.9/86.4         & 72.5/95.5      & 81.7/97.6 \\
    IS FT         & 93.5/99.6     & 62.1/91.4         & 60.1/91.2      & 99.1/100.0 \\
    \bottomrule
    \end{tabular}
    \vspace{1em}
    \caption{\textbf{Performance on datasets for generalization and adaptation across sessions, or intersession performance.} In this case, the test set is the second session of a left-out subject. FT stands for fine-tuning. IS FT stands for intersession fine tuning. Before fine-tuning, the model is trained on all subjects other than the left-out subject. Fine-tuning involves training on the left-out subject's first session.}
    \label{tab:intersession}
\end{table}

\begin{table}[h]
    \centering
    {
    \small
    \begin{tabular}{lccccccccc}
        \toprule
        \textbf{Task}     & \bf{UCI EMG}   & \bf{Ninapro DB5}  & \bf{Ninapro DB2} & \bf{Ninapro DB3} & \bf{MCS}   \\
        \midrule
        \multicolumn{6}{l}{\textbf{Varying proportions, transitions classified, finetuning}} \\
        \toprule
        FT-20\%           &    93.0/99.1   &        59.6/91.0  & 47.6/84.4        & 45.5/80.8        & 85.3/97.6   \\
        FT-40\%           &    93.6/99.5   &        66.5/93.5  & 48.2/84.7        & 45.8/80.9        & 88.0/98.5   \\
        FT-60\%           &    95.0/99.5   &        72.5/95.4  & 46.6/84.1        & 44.2/79.5        & 91.0/99.1   \\
        FT-80\%           &    98.0/99.9   &        65.0/93.1  & 44.7/82.5        & 40.3/78.2        & 89.0/98.7   \\
        \bottomrule
    \end{tabular}
    }
    \vspace{1em}
    \caption{\textbf{Performance when including transition data.} This table shows performance when including transition data and not only during isometric holds of a gesture. The label on the entire window of data is set as the label of the data at the last time step of the window.}
    \label{tab:transition-tasks}
\end{table}

\subsection{Domain Generalization Algorithms}

We test additional training algorithms, which have been tested in previous work on domain generalization, to compare how well they work compared to standard supervised learning~\cite{gulrajani2020search, koh2021wilds}. In these experiments, we again use the best performing preprocessing method and model architecture for the dataset, but include the use of invariant risk minimization (IRM), and correlation alignment (CORAL)~\cite{gulrajani2020search, koh2021wilds}. We test these algorithms that take into account the different domains during training: 1) IRM minimizes a loss that penalizes models where the optimal classifier differs across domains, and 2) CORAL aligns the covariances of the feature representations between domains through a loss term. The resulting generalization and few-shot fine-tuning results are shown in Appendix \ref{appendix:domain-generalization-techniques}. Overall performance is comparable to with training using standard cross-entropy loss, which is a similar result found in \citet{gulrajani2020search} and \citet{koh2021wilds}.

\section{Limitations and Future Work}
\label{section:limitations-and-future-work}

The present benchmarking studies in this work have not evaluated EMG classification performance with activity maps exceeding a resolution of 224x224 pixels. Higher resolution activity maps may enhance inter-subject performance when employing spectrogram or continuous wavelet transform (CWT) image preprocessing techniques, particularly for datasets with a greater number of electrodes. This is due to the potential reduction in downsampling of the resulting time-frequency transformed images for each electrode, thereby preserving specific features within the transformed data.

Among the datasets analyzed, as presented in Table \ref{tab:datasets-vs-models}, the FlexWear-HD dataset achieved the highest leave-one-subject-out cross-validation accuracy, registering an 83.9\% test accuracy for 10-class classification. While robust generalization to new subjects for EMG control interfaces remains an ongoing research challenge \cite{ctrl2024generic}, future work, leveraging larger datasets and models, suggests the potential for developing a universal EMG classifier. Such a classifier would exhibit the ability to generalize or adapt rapidly to EMG data from novel subjects, thereby facilitating the robust deployment of EMG-based control interfaces.

Datasets such as CapgMyo, Hyser, Myo Dataset, and FlexWear-HD do not present transition data between gestures, likely due to the added complexity of labeling short dynamic transition data for a gesture-based classification interface. However, we note that in previous robot control work with the FlexWear-HD~\cite{yang2023high}, while these transition windows were also not taken into account during training, users were able to control a robot using streams of EMG data classified in real-time to perform complex assistive tasks through teleoperation. However, for the datasets that do include transition data, such as the Ninapro datasets, MCS, and UCI EMG, we include these results in Table \ref{tab:transition-tasks} where the transition windows are also classified. 

Currently, none of these datasets capture gesture data performed in real-world scenarios where users engage in everyday activities or utilize an EMG-based gesture classification system for interacting with computers or robots. Future work would benefit from the development of a dataset that includes ground truth data on gestures performed spontaneously by the user, outside of controlled environments and cue-based systems. Such a dataset would capture differences on how gestures are performed in real-world settings, improving the performance and adaptability of EMG-based gesture classification systems once properly benchmarked.

\section{Conclusion}

This study introduces a new tool for benchmarking EMG datasets, providing insights into the effectiveness of various classification methods, EMG preprocessing techniques, potential for real-world applications, and open research problems in learning-based EMG gesture classification for the research community. The performance of these methods shows promising results for enhancing the generalization of EMG-based systems, especially through a standardized format to compare performances metrics. While benchmarking, we find that adaptation using data from the validation subject can significantly enhance performance, requiring only a small amount of data from the subject.

\section*{Acknowledgments}
This material is based upon work supported by the National Science Foundation under Grant No. 2341352 and Grant No. DGE2140739.

% Use unnumbered first level headings for the acknowledgments. All acknowledgments
% More information about this disclosure can be found at: \url{https://neurips.cc/Conferences/2024/PaperInformation/FundingDisclosure}.

% Do {\bf not} include this section in the anonymized submission, only in the final paper. You can use the \texttt{ack} environment provided in the style file to automatically hide this section in the anonymized submission.
% \end{ack}

% \section*{References}

% References follow the acknowledgments in the camera-ready paper. Use unnumbered first-level heading for
% the references. Any choice of citation style is acceptable as long as you are
% consistent. It is permissible to reduce the font size to \verb+small+ (9 point)
% when listing the references.
% Note that the Reference section does not count towards the page limit.
% \medskip

\bibliography{references}

%%%%%%%%%%%%%%%%%%%%%%%%%%%%%%%%%%%%%%%%%%%%%%%%%%%%%%%%%%%%

\appendix

\section{Appendix / supplemental material}

\subsection{EMG Signals}
\label{appendix:emg}

EMG sensors detect voltage signals due to motor neurons activating the attempted contraction of muscle fibers. Muscle contraction within the body is caused by the activation of motor neurons that innervate a set of around 100 to 1000 muscle fibers. This set of innervated muscle fibers is called a motor unit~\cite{purves2001neuroscience}. When a neuron's voltage spike propagates to the end of its axon, the neuron releases acetylcholine neurotransmitters into the neuromuscular junction, activating acetylcholine-modulated ion channels that allow sodium ions to enter the muscle cell. The influx of sodium ions causes voltage-modulated ion channels for sodium to open as well, propagating the voltage spike from the action potential along the muscle fiber~\cite{mukund2020skeletal}.

Once the sodium voltage spike propagates along the muscle fiber membrane and reaches an area of the membrane called the T-tubules, a voltage-sensitive reaction is triggered, causing calcium ion channels in the sarcoplasmic reticulum organelle to open. When calcium ions move into the cytoplasm of the muscle fiber, they bind to troponin proteins within the muscle fiber, exposing binding sites on actin filaments. Myosin heads bind to these sites and pull actin filaments toward the middle of a sarcomere, causing muscular contraction~\cite{mukund2020skeletal}.

Because of the speed of the propagation of the action potential in a muscle fiber and the superposition of voltage signals of a population of motor units firing at once~\cite{preston2005basic}, examinations involving surface EMG will see signals of interest from EMG activity ranging from 10 Hz to 500 Hz~\cite{nazmi2016review}. This distribution of EMG activity over frequency is dependent on both impedances in the body between the electrode and the muscle fibers, and on the proportion of slow-twitch (Type I) and fast-twitch (Type II) muscle fibers in the anatomy~\cite{mukund2020skeletal}. The spatial distribution of EMG activity around the arm further distinguishes the specific intended gesture of the subject~\cite{preston2005basic}. For example, the activation of more muscle fibers on the palmar side of the forearm (specifically the flexor carpi muscles) may distinguish wrist flexion from other gestures. The activation of different proportions of slow and fast muscle fibers as well as the spatial distributions of muscle activation change over time depending on the amount of fatigue and amount of effort attempted by the user, for example due to muscular compensation~\cite{chua2024analysis, liu2021muscle}. By detecting voltage signals from patterns of muscle voltage spikes over time and spatially around the forearm, we are able to decipher high-dimensional motor intent from as many as 20 muscles around the forearm. 

\subsection{EMG Devices}

A common EMG device, used by 3 of the datasets in our benchmark, is the Myo Armband. The Ninapro DB5, Myo Dataset, and the UCI EMG dataset all use this device. Each armband contains 8 stainless steel electrodes. Due to the use of dry stainless steel electrodes without the use of electrolyte gel and in order to save energy for a battery-powered wireless wearable device, respectively, these devices have relatively high electrode-skin impedances~\cite{yang2022insight} and relatively low sampling rates (200 Hz).

Other common devices used for EMG measurements are the individually-placed bipolar Delsys Trigno sensors, which use dry silver electrodes~\cite{atzori2012building}; high-density EMG arrays, which use gels on top of exposed metal pads on flexible printed circuit boards~\cite{geng2016gesture}; as well as individually-placed sticky hydrogel electrodes, which are often commonly used in electrocardiograms~\cite{ozdemir2022dataset}. 

\subsection{EMG Preprocessing into Heatmaps}

\label{appendix:preprocessing}

Preprocessing EMG time-series data as heatmap images and classifying them using CNNs has yielded high classification accuracy results in prior work~\cite{geng2016gesture, ozdemir2022hand}, setting the state-of-the-art for gesture classification in 2016~\cite{geng2016gesture} for some popular EMG datasets, including Ninapro DB1, Ninapro DB2, and the CSL-HDEMG. After this work, many papers used similar heatmap image preprocessing methods, particularly by transforming raw data into heatmaps~\cite{hu2018novel}, first doing feature extraction methods such as root-mean-square windows before transforming into heatmaps~\cite{yang2023high}, or first transforming the images into time-frequency plots (such as spectrograms, or continuous wavelet transforms)~\cite{ozdemir2022hand}. 

\subsection{Dataset Details in Benchmark}

\begin{table}[h!]
    \centering
    {
    \tiny
    \begin{tabular}{@{}llccccccc@{}}
    \toprule
    \textbf{Dataset} & \textbf{Channels} & \textbf{Subjects} & \textbf{Gestures} & \textbf{Sample Len (ms)} & \textbf{Step Len (ms)} & \textbf{Sampling Freq (Hz)} & \textbf{Sessions} & \textbf{Samples} \\ 
    \midrule
    Myo Dataset     & 8 dry           & 18       & 7        & 250              & 50               & 200                  & 1          & 48030   \\
    UCI EMG         & 8 dry           & 36       & 6        & 250              & 50               & 1000                 & 2          & 12870  \\
    NinaproDB5      & 16 dry          & 10       & 10       & 250              & 50               & 200                  & 1          & 39597   \\
    Capgmyo         & 128 gelled      & 10       & 8        & 250              & 50               & 1000                 & 2          & 25600   \\
    NinaproDB2      & 12 dry bipolar  & 40       & 10       & 250              & 250              & 2000                 & 1          & 38810   \\
    NinaproDB3      & 12 dry bipolar  & 11       & 10       & 250              & 50               & 2000                 & 1          & 64426   \\
    MCS             & 4 gelled bipolar & 40      & 7        & 250              & 250              & 2000                 & 1          & 28000   \\
    Hyser           & 256 gelled      & 20       & 10       & 250              & 125              & 2048                 & 2          & 8239    \\
    FlexWear-HD     & 64 hydrogel     & 13       & 10       & 250              & 50               & 4000                 & 2          & 46116   \\
    \bottomrule
    \end{tabular}
    }
    \vspace{1em}
    \caption{Summary of information on EMG datasets. The number of sample lengths, and step lengths were prescribed in this study. The number of gestures used is also less than all that is available for the UCI EMG, Hyser, NinaproDB2, NinaproDB5, and MCS datasets similar to other studies that tested generalization~\cite{lu2022domain, ozdemir2022hand, li2023deep}. The number of reported samples are from session 1, as all studies other than those in Table \ref{tab:intersession} use only the first session. }
    \label{tab:emg_datasets}
\end{table}

In Table \ref{tab:emg_datasets}, we show details about the datasets in this benchmark in a condensed table format. For the UCI EMG, Hyser, NinaproDB2, NinaproDB5, and MCS datasets, we use less than the total number of gestures available, similar to other studies that have tested generalization; we select these gestures based on papers that have previously tested with these datasets before and that are based on basic wrist and finger movements~\cite{lu2022domain, ozdemir2022hand, li2023deep}. For most datasets, we use a step length of 50 ms, although for some datasets with a high number of electrodes and sampling rate, we increase the step length to make computation more tractable.

\begin{table}[h!]
    \centering
    {
    \tiny
    \begin{tabular}{@{}lcccccccccc@{}}
    \toprule
    \textbf{Dataset} & \textbf{Rest} & \textbf{Radial} & \textbf{Flexion} & \textbf{Ulnar} & \textbf{Extension} & \textbf{Fist} & \textbf{Abduction} & \textbf{Adduction} & \textbf{Supination} & \textbf{Pronation} \\ 
    \midrule
    Myo Dataset & 6861 & 6861 & 6855 & 6864 & 6863 & 6865 & 6861 & & &  \\
    UCI EMG & 4286 & 4325 & 4273 & 4347 & 4310 & 4140 & & & &  \\
    NinaproDB5 & 4102 & 3335 & 4199 & 4756 & 3709 & 3614 & 3545 & 3926 & 4326 & 4085  \\
    Capgmyo &  &  &  &  &  & 1600 & 1600 & 1600 & &   \\
    NinaproDB2 & 4183 & 3971 & 3149 & 3863 & 3091 & 3555 & 4094 & 4090 & 4456 & 4358   \\
    NinaproDB3 & 6556 & 6511 & 6123 & 6341 & 6085 & 6206 & 6477 & 7075 & 7245 & 5807   \\
    MCS & 4000 & 4000 & 4000 & 4000 & 4000 & 4000 & 4000 & & &   \\
    Hyser &  & 1582 & 1680 & 1673 & 1666 & 1680 & 1680 & & 1638 & 1638   \\
    FlexWear-HD & 4644 & 4572 & 4572 & 4572 & 4680 & 4644 & 4644 & & 4644 & 4644   \\
    \bottomrule
    \end{tabular}
    }
    \vspace{1em}
    \caption{The number of samples for common gestures in each EMG dataset. The full gesture names are Radial and Ulnar Deviation; Finger Abduction and Adduction; Wrist Flexion, Extension, Supination, and Pronation.}
    \label{tab:common_gestures}
\end{table}

We specify the number of samples for each gesture in Tables \ref{tab:common_gestures} and \ref{tab:gesture-samples}. All samples were created from overlapping 250 ms windows of the raw data, where each dataset's step length determined the amount of time between the start of one window and the next. 250 ms was selected as the standard sample length since longer window lengths generally improve classification but a delay of over 250 ms would make the classifier unsuitable for real-time usage ~\cite{smith2010determining}. Windows at the transition between gestures were excluded, ensuring that every window corresponds to a single gesture.

To prevent large class imbalance in Ninapro DB5, the rest gesture was subsampled to the average number of samples of the other gestures. The variation in sample numbers across a dataset is due to small differences in repetition length during data collection and incorrectly performed repetitions being withheld from the dataset.

\setlength{\tabcolsep}{4pt}
\begin{table}[h!]
    \centering
    {
    \tiny
    \begin{tabular}{@{}llccccccc@{}}
    \toprule
    \textbf{Dataset} & \textbf{Gesture Type} & \textbf{Thumb, Index} & \textbf{Thumb, Middle} & \textbf{Thumb, Index, Middle} & \textbf{Thumb} & \textbf{Index} & \textbf{Index, Middle} & \textbf{All But Thumb} \\ 
    \midrule
    Capgmyo     & Extension & --   & --   & 1600 & 1600 & 1600 & 1600 & 1600 \\
    Hyser       & Pinch     & 1673 & 1603 & --   & --   & --   & --   & --   \\
    FlexWear-HD & Pinch     & --   & --   & 4500 & --   & --   & --   & --   \\
    \bottomrule
    \end{tabular}
    }
    \vspace{1em}
    \caption{The number of samples for less common gestures in each EMG dataset. The column names refer to the fingers involved in each gesture.}
    \label{tab:gesture-samples}
\end{table}

\setlength{\tabcolsep}{4pt}
\begin{table}[h!]
    \centering
    {
    \tiny
    \begin{tabular}{@{}lccccccccc@{}}
    \toprule
    \textbf{} & \textbf{Myo Dataset} & \textbf{UCI EMG} & \textbf{NinaproDB5} & \textbf{Capgmyo} & \textbf{NinaproDB2} & \textbf{NinaproDB3} & \textbf{MCS} & \textbf{Hyser} & \textbf{FlexWear-HD} \\ 
    \midrule
    Time per Repetition        & 5 & 3 & 5 & 3  & 5 & 5 & 6 & 1 & 2 \\
    Inter-Repetition Rest Time & 5 & 3 & 3 & 7  & 3 & 3 & 4 & 2 & 2.5\\
    Repetitions per Gesture    & 4 & 1/1 & 6 & 10/10 & 6 & 6 & 5 & 3/3 & 10/5\\
    \bottomrule
    \end{tabular}
    }
    \vspace{1em}
    \caption{The number of seconds each repetition of a gesture was performed for, the number of seconds participants rested between repetitions, and the number of repetitions that were performed for each gesture. A / separates the repetitions in the first session from the second for datasets with multiple sessions.}
    \label{tab:gesture-times}
\end{table}

\begin{table}[h!]
    \centering
    {
    \tiny
    \begin{tabular}{@{}ccccccc@{}}
    \toprule
    \textbf{Reference Study} & \textbf{Chronological Split} & \textbf{Random Split} & \textbf{Mixed Split} & \textbf{Inter-Subject} & \textbf{Inter-Session} & \textbf{Inter-Trial} \\ 
    \midrule
    ~\citet{zhang2022research}     & \redx & \redx & \greencheck & \redx & \redx & \greencheck \\
    ~\citet{lin2022reliability}    & \redx & \redx & \greencheck & \redx & \redx & \greencheck \\
    ~\citet{lee2021electromyogram} & \redx & \greencheck & \redx & \redx & \redx & \redx \\
    ~\citet{tuncer2022novel}       & \redx & \greencheck & \redx & \redx & \redx & \redx \\
    ~\citet{azhiri2021emg}         & \greencheck & \redx & \redx & \redx & \redx & \greencheck \\
    ~\citet{sri2021classification} & \redx & \redx & \greencheck & \redx & \redx & \greencheck \\
    ~\citet{fatimah2021hand}       & \redx & \greencheck & \redx & \redx & \redx & \redx \\
    ~\citet{chen2021surface}       & \redx & \redx & \greencheck & \redx & \redx & \greencheck \\
    ~\citet{rahimian2021fs}        & \redx & \redx & \greencheck & \redx & \redx & \greencheck \\
    ~\citet{kim2021semg}           & \redx & \redx & \greencheck & \redx & \greencheck & \redx \\
    ~\citet{fu2021finger}          & \redx & \greencheck & \redx & \redx & \redx & \redx \\
    ~\citet{abbaspour2020evaluation} & \redx & \greencheck & \redx & \redx & \redx & \redx\\
    ~\citet{gautam2020locomo}      & \redx & \greencheck & \redx & \redx & \redx & \redx\\
    ~\citet{ozdemir2020emg}        & \redx & \greencheck & \redx & \redx & \redx & \redx\\
    ~\citet{he2020biometric}       & \redx & \greencheck & \redx & \redx & \redx & \redx\\
    ~\citet{shen2019movements}     & \greencheck & \redx & \redx & \redx & \redx & \greencheck\\
    ~\citet{qi2019intelligent}     & \greencheck & \redx & \redx & \redx & \greencheck & \redx\\
    ~\citet{bhagwat2020electromyogram} & \redx & \greencheck & \redx & \redx & \redx & \redx\\
    ~\citet{kim2019development}    & \redx & \redx & \greencheck & \redx & \greencheck & \redx\\
    ~\citet{purushothaman2018identification} & \redx & \greencheck & \redx & \redx & \redx & \redx\\
    ~\citet{saikia2018combination} & \redx & \greencheck & \redx & \redx & \redx & \redx\\
    ~\citet{baldacchino2018simultaneous} & \redx & \redx & \greencheck & \redx & \redx & \greencheck\\
    ~\citet{phukpattaranont2018evaluation} & \redx & \greencheck & \redx & \redx & \redx & \redx\\
    ~\citet{sezgin2019new}         & \redx & \greencheck & \redx & \redx & \redx & \redx \\
    ~\citet{jiralerspong2017experimental} & \greencheck & \redx & \redx & \redx & \greencheck & \redx\\
    ~\citet{rahimi2016hyperdimensional} & \redx & \greencheck & \redx & \redx & \redx & \redx \\
    ~\citet{ariyanto2015finger}    & \redx & \redx & \greencheck & \redx & \redx & \greencheck\\
    ~\citet{gijsberts2014movement} & \redx & \redx & \greencheck & \redx & \redx & \greencheck\\
    ~\citet{al2013classification}  & \redx & \redx & \greencheck & \redx & \redx & \greencheck\\
    ~\citet{khushaba2012toward}    & \redx & \redx & \greencheck & \redx & \redx & \greencheck\\
    ~\citet{tang2012hand}          & \redx & \redx & \greencheck & \redx & \redx & \greencheck\\
    ~\citet{khushaba2011electromyogram} & \redx & \redx & \greencheck & \redx & \redx & \greencheck\\
    ~\citet{naik2012identification} & \redx & \redx & \greencheck & \redx & \redx & \greencheck\\
    \bottomrule
    \end{tabular}
    }
    \vspace{1em}
    \caption{This table shows the ways in which each reference study splits its data into train and test sets. Chronological split has the test set as the last portion of the dataset to be collected. Random split has a random subset of the dataset as the test set; k-fold CV is included here. Mixed split has the test set consisting of distinct subjects, sessions, or trials not collected last chronologically. Inter-subject, inter-session, and inter-trial refer to the dataset attribute that chronological splits and mixed splits partitioned across.}
    \label{tab:data-splits}
\end{table}

In this table we show the data splitting methods used by all 33 EMG gesture classification studies reviewed in ~\cite{sultana2023systematic}. Studies were systematically selected based on several criteria including being recent and published after 2010, being open access, and having finger gestures within the set of gestures for classification. We note that data splitting methods used in our study (chronological splitting, inter-subject, and inter-session) are much less common in the literature, but are more indicative of capacity for generalization than more commonly used methods. Notably, inter-session testing leads to more distribution shift than inter-trial since it usually involves the subject removing and reattaching the measurement device after a significant rest period; inter-subject testing has even greater distribution shift.

\subsection{Learning-based generalization or adaptation prior work}

\label{appendix:generalization-prior-work}

A small proportion of EMG classification papers use splits based on leave-one-subject-out (LOSO) cross validation, despite its importance for real-world deployment. None of the papers reviewed in the EMG classification review by \citet{sultana2023systematic} evaluate using LOSO-CV or other intersubject accuracy tests. Among the works that do test on intersubject accuracy, \citet{ozdemir2022hand} presents a fine-tuned ResNet50 model on a 4-electrode dataset, achieving a 94.41\% LOSO-CV accuracy for 7 gestures. Similarly, \citet{li2023deep} reports CNN models that attain an 85.4\% LOSO-CV accuracy using a dataset they collected with 256 electrodes for 10 gestures, although the published dataset uses different subjects. Additionally, works by \citet{wang2023pruning}, \citet{zhang2023multi}, and \citet{xu2023cross} on their own unpublished datasets achieve LOSO-CV accuracies of 80.3\% for 5 gestures, 73\% for 6 gestures, and 60.8\% for 17 gestures, respectively.

In the literature, among the prior work that tests on leaving at least one subject out for a pre-existing EMG dataset published by another group, only one study has achieved greater than 50\% accuracy. \citet{lu2022domain} achieves an average accuracy of 77\% for 6 gestures. Instead of the typical LOSO cross validation, \citet{lu2022domain} uses a train-test split, randomly leaving nine out of thirty-six subjects as the test set, and utilizes an EMG dataset published by \citet{krilova2019emg}. Other studies by \citet{islam2024surface}, \citet{du2017surface}, and \citet{wei2021hierarchical} that employ LOSO cross validation on previously available EMG datasets, such as NinaPro or CapgMyo, achieve approximately or less than 50\% accuracy, with \citet{islam2024surface} and \citet{wei2021hierarchical} not reporting their exact accuracies due to low performance.

There is a larger body of prior work that test on out-of-domain adaptation tasks, involving using small amounts of data, either labeled, unlabeled, or both, from the subject that the model is tested on. For example, work from \citet{cote2019deep}, \citet{li2023deep}, \citet{zhang2023multi}, \citet{islam2024surface}, \citet{du2017surface}, and \citet{wei2021hierarchical} all evaluate adaptation. All of this work varies in how much data is used, whether unlabeled or labeled data is used from the validation subject, and the datasets tested. A common version of these out-of-domain adaptation tasks involves \textbf{intersession accuracy tests (IAT)}, in which an entire session of data collection is left-out as the test set. This approach is practical for predicting gestures in a new session for a subject with data from a previous session. The test session can range from data collected on the same day~\cite{yang2023high, krilova2019emg} to several days after the initial session~\cite{du2017surface, li2023deep}, and it varies based on whether the sensors were re-donned for the test session~\cite{du2017surface, li2023deep} or continuously worn since the previous data collection~\cite{yang2023high}.

\subsection{Classification Metrics}
\label{appendix:classification-metrics}
For all tests, we split the data into training, validation, and test sets. The training set includes data from all subjects except the left-out subject. The validation and test sets are evenly split subsets of the left-out subject's data, with the validation set being the earlier subset and the test set being the later subset in time. The test set is only evaluated at the final epoch. When evaluating task 2, we use a fine-tuning set that consists of an initial subset of data from the evaluation subject.

We report two classification metrics: average test classification accuracy and average area under the receiver operating characteristic (AUROC) for the test set in an average one-vs-rest split. Since our datasets are balanced, as shown in Table \ref{tab:gesture-samples}, the average test classification accuracy is unbiased towards the performance of any specific class. The AUROC is also commonly used to evaluate the classifier's discrimination ability.

\subsection{Additional Details on Datasets and Benchmarking}
\label{appendix:additional-details-on-datasets}

Although there are several published EMG datasets, there are several variables that can differ from dataset to dataset, for example: number of electrodes, electrode materials, recording hardware involving differing hardware amplifiers, hardware filters, and analog-to-digital converters, numbers of participants, sets of gestures, and sampling rates~\cite{atzori2014electromyography, du2017surface, cote2019deep, ozdemir2022dataset, krilova2019emg, jiang2021open}. Because there is no standard set of hardware or gestures that people have used for studies involving EMG gesture classification, it is important to find learning-based models that perform well across these different variables.

\subsubsection{Ninapro}
From prior work, we note that \citet{wei2021hierarchical} mentioned low leave-one-subject-out cross validation accuracy, reporting around 30\% in preliminary experiments on the Ninapro DB1 dataset. This may be due to the difficulty required in precise individual sensor placements around the arm for use in gesture classification for the Ninapro DB1 dataset~\cite{atzori2012building}. In addition, an assessment of the Ninapro datasets by~\citet{chang2020assessment} of DB2 to DB8 have found low signal to noise ratios for some sub-datasets, as well as occurrences of mislabeling across all these sub-datasets. The numbers of subjects, number of gestures, and the types of sensor can vary between the datasets~\cite{atzori2012building, atzori2014electromyography, pizzolato2017comparison}, although the range of number of electrodes varies from only 10 to 16 electrodes. 

\subsubsection{CapgMyo}
From prior work by \citet{du2017surface}, LOSO classification accuracy was reported to be 39.0\%, although this rises to 55.3\% using an adaptive batch normalization method when including data from the left-out subject. We note that the device used for measuring EMG seems to require the placement of 8 individual modules on the arm, which may cause significant variations required for individual placements between sessions and subjects. The minimum number of days between recording sessions is 7 days for the same subject. 

\subsubsection{Myo Dataset}
Prior work from \citet{cote2019deep} and \citet{lin2020normalisation} classifying the Myo dataset focuses on domain adaptation, using some data from the validation subject during training. These works show 98.31\% and 94.53\% classification accuracy, respectively. Although the former work uses the full 36 subject dataset, the latter exclusively uses the 17 subjects designated as the "evaluation dataset". Our benchmark also uses the evaluation dataset, which now includes 18 subjects. For this dataset, there are additional details on the placement of the Myo Armband on the arm, namely that orientation of the Myo Armband is placed such that the light on the armband faces the hand, the tightness of the armband is configured to maximum tightness, and the armband is slid onto the arm until the inner circumference of the armband matches the forearm~\cite{cote2019deep}. Due to this placement procedure, based on figures included in \cite{cote2019deep}, the Myo Armband is placed anywhere from the thickest part of the forearm (close to the elbow), to a few centimeters from the wrist. 

\subsubsection{UCI EMG}
Although there is not a significant amount of detail published with this dataset, it is specified that the Myo Armband device is used, and bluetooth is used to send to a PC from the device. In a test involving testing on left-out-subjects in \citet{lu2022domain}, an average validation accuracy of 77\% is achieved while leaving 9 out of 36 subjects out as the validation set at a time. We note that although the raw data is recorded at 1000 Hz while the Myo Armband can only record at 200 Hz, the authors seem to have upsampled the data. This can be seen in the raw data from the many repeated values across timesteps. We note that for Table \ref{tab:adaptation-tasks}, the UCI EMG dataset is fine-tuned on the first X\% of samples after concatenating samples from both sessions together, where each session only has 1 gesture repetition

\subsubsection{Hyser Dataset}
Similar to work from \citet{li2023deep}, which achieves a LOSO-CV accuracy of 85.4\% and a leave-one-session-out accuracy of 82.2\% for a dataset collected using the same protocol as the Hyser dataset, in our benchmarking script, we also use only 10 of the 34 gestures for a more tractable learning problem for real-world use. We note we were not able to achieve the same LOSO-CV accuracy, although the subjects in the published dataset are different than the subjects in \citet{li2023deep}. We note that because Hyser subject 5 seems to only have the first 9 out of 10 gestures we classify for in their first session, we ignore the classification of Hyser subject 5 for the results tables we generate because we want to keep a consistent number of gestures between subjects for classification. Data collection sessions are on separate days for the same subject. 

\subsubsection{FlexWear-HD Dataset}
\label{appendix:flexwear-hd}
The FlexWear-HD dataset includes 13 subjects, who perform 15 total repetitions of 10 gestures. An easy-to-wear, reusable high density 64-electrode hydrogel array is used, with the device wrapped around the proximal forearm with a Velcro strip. The strip is placed approximately in the same orientation across subjects with palpation of the location of the ulnar bone as a landmark to place the electrode device. Additional detail on the device and placement is found in \cite{yang2023high}. Data from two sessions for this dataset is provided, with the second session occurring about one hour after the first session. By performing well on leave-one-session-out tests on this dataset, we can evaluate robustness of the wearable EMG device as a control interface over timescales of around an hour after EMG control interface use. 
Before inclusion in the FlexWear-HD dataset, participants gave their written informed consent and agreed that this material can be used in journals and other public media. The study protocol was approved by the Carnegie Mellon University Institutional Review Board, protocol 2021.00000121.

\subsection{Additional Results Classifying Phase-Based Information}
\label{appendix:phase-preprocessing}

We anticipated that phase-based featurization from each electrode's data may be able to be used to capture the spatial propagation of muscle action potentials. In order to test this, we extracted the phase from STFT and the instantaneous phase from the intrinsic mode functions of a Hilbert transform for each electrode. Results are shown in Table \ref{tab:phase}.

\begin{table}[h]
    \centering
    \small
    \begin{tabular}{lc}
    \toprule
    \textbf{Task} & \\
    \midrule
    \multicolumn{2}{c}{\hspace{-1.5em}\textbf{Pretraining before few-shot fine-tuning, LOSO-CV}} \\
    \midrule
    Phase Spectrogram     & 22.4/64.3      \\
    HHT Phase & 25.1/67.1 \\
    \midrule
    \multicolumn{2}{c}{\textbf{Finetuning with first 20\% of data from left-out subject}} \\
    \midrule
    Phase Spectrogram     & 27.0/69.3      \\
    HHT Phase & 31.0/72.3 \\
    \bottomrule
    \end{tabular}
    \vspace{1em}
    \caption{Performance using phase-based representations for the Ninapro-DB5 dataset.}
    \label{tab:phase}
\end{table}

\subsection{Additional Results Using Domain Generalization Techniques}
\label{appendix:domain-generalization-techniques}

We tested typical domain generalization techniques, such as invariant risk minimization (IRM), and correlation alignment (CORAL). These techniques were tested in two previous benchmarks on domain generalization~\cite{koh2021wilds, gulrajani2020search}. The results for the Ninapro DB5 dataset is in Table \ref{tab:dom-gen}.

\begin{table}[h]
    \centering
    \small
    \begin{tabular}{lc}
    \toprule
    \textbf{Task} & \\
    \midrule
    \multicolumn{2}{c}{\hspace{-1.5em}\textbf{Pretraining before few-shot fine-tuning, LOSO-CV}} \\
    \midrule
    CORAL       & 41.0/78.6      \\
    IRM         & 42.3/78.7      \\
    \midrule
    \multicolumn{2}{c}{\textbf{Finetuning with first 20\% of data from left-out subject}} \\
    \midrule
    CORAL       & 68.8/93.9      \\
    IRM         & 68.4/93.5      \\
    \bottomrule
    \end{tabular}
    \vspace{1em}
    \caption{Performance using domain generalization-based training methods for the Ninapro-DB5 dataset.}
    \label{tab:dom-gen}
\end{table}

%%%%%%%%%%%%%%%%%%%%%%%%%%%%%%%%%%%%%%%%%%%%%%%%%%%%%%%%%%%%

\clearpage

\section*{NeurIPS Paper Checklist}

\begin{enumerate}

\item {\bf Claims}
    \item[] Question: Do the main claims made in the abstract and introduction accurately reflect the paper's contributions and scope?
    \item[] Answer: \answerYes{} % Replace by \answerYes{}, \answerNo{}, or \answerNA{}.
    \item[] Justification: The three main claims includes 1) a released codebase which we link as a GitHub repo that is made public for others to replicate code and run benchmarking \texttt{github.com/jehanyang/emgbench}, 2) presentation of a dataset using an easy-to-wear high-density EMG sensor that we link on our website \texttt{emgbench.github.io}, and 3) benchmarking results across tasks, which we present in Section \ref{section:results}. 
    \item[] Guidelines:
    \begin{itemize}
        \item The answer NA means that the abstract and introduction do not include the claims made in the paper.
        \item The abstract and/or introduction should clearly state the claims made, including the contributions made in the paper and important assumptions and limitations. A No or NA answer to this question will not be perceived well by the reviewers. 
        \item The claims made should match theoretical and experimental results, and reflect how much the results can be expected to generalize to other settings. 
        \item It is fine to include aspirational goals as motivation as long as it is clear that these goals are not attained by the paper. 
    \end{itemize}

\item {\bf Limitations}
    \item[] Question: Does the paper discuss the limitations of the work performed by the authors?
    \item[] Answer: \answerYes{} % Replace by \answerYes{}, \answerNo{}, or \answerNA{}.
    \item[] Justification: Limitations are listed in Section \ref{section:limitations-and-future-work}. Specifically, limitations include testing with only a single seed for each experiment due to the amount of compute required for training and testing individual models for each subject. Limitations also include the use of a limited 224x224 heatmap resolution, which may lose details after preprocessing time-series data into spectrograms and CWTs that are tiled into a single image. 
    \item[] Guidelines:
    \begin{itemize}
        \item The answer NA means that the paper has no limitation while the answer No means that the paper has limitations, but those are not discussed in the paper. 
        \item The authors are encouraged to create a separate "Limitations" section in their paper.
        \item The paper should point out any strong assumptions and how robust the results are to violations of these assumptions (e.g., independence assumptions, noiseless settings, model well-specification, asymptotic approximations only holding locally). The authors should reflect on how these assumptions might be violated in practice and what the implications would be.
        \item The authors should reflect on the scope of the claims made, e.g., if the approach was only tested on a few datasets or with a few runs. In general, empirical results often depend on implicit assumptions, which should be articulated.
        \item The authors should reflect on the factors that influence the performance of the approach. For example, a facial recognition algorithm may perform poorly when image resolution is low or images are taken in low lighting. Or a speech-to-text system might not be used reliably to provide closed captions for online lectures because it fails to handle technical jargon.
        \item The authors should discuss the computational efficiency of the proposed algorithms and how they scale with dataset size.
        \item If applicable, the authors should discuss possible limitations of their approach to address problems of privacy and fairness.
        \item While the authors might fear that complete honesty about limitations might be used by reviewers as grounds for rejection, a worse outcome might be that reviewers discover limitations that aren't acknowledged in the paper. The authors should use their best judgment and recognize that individual actions in favor of transparency play an important role in developing norms that preserve the integrity of the community. Reviewers will be specifically instructed to not penalize honesty concerning limitations.
    \end{itemize}

\item {\bf Theory Assumptions and Proofs}
    \item[] Question: For each theoretical result, does the paper provide the full set of assumptions and a complete (and correct) proof?
    \item[] Answer: \answerNA{} % Replace by \answerYes{}, \answerNo{}, or \answerNA{}.
    \item[] Justification: We do not present theoretical results. 
    \item[] Guidelines:
    \begin{itemize}
        \item The answer NA means that the paper does not include theoretical results. 
        \item All the theorems, formulas, and proofs in the paper should be numbered and cross-referenced.
        \item All assumptions should be clearly stated or referenced in the statement of any theorems.
        \item The proofs can either appear in the main paper or the supplemental material, but if they appear in the supplemental material, the authors are encouraged to provide a short proof sketch to provide intuition. 
        \item Inversely, any informal proof provided in the core of the paper should be complemented by formal proofs provided in appendix or supplemental material.
        \item Theorems and Lemmas that the proof relies upon should be properly referenced. 
    \end{itemize}

    \item {\bf Experimental Result Reproducibility}
    \item[] Question: Does the paper fully disclose all the information needed to reproduce the main experimental results of the paper to the extent that it affects the main claims and/or conclusions of the paper (regardless of whether the code and data are provided or not)?
    \item[] Answer: \answerYes{} % Replace by \answerYes{}, \answerNo{}, or \answerNA{}.
    \item[] Justification: We provide a benchmarking codebase that makes it easy to reproduce our results at \texttt{github.com/jehanyang/emgbench}, with clear instructions in a \texttt{README.md} document and corresponding configuration files for the results produced in our tables. 
    \item[] Guidelines:
    \begin{itemize}
        \item The answer NA means that the paper does not include experiments.
        \item If the paper includes experiments, a No answer to this question will not be perceived well by the reviewers: Making the paper reproducible is important, regardless of whether the code and data are provided or not.
        \item If the contribution is a dataset and/or model, the authors should describe the steps taken to make their results reproducible or verifiable. 
        \item Depending on the contribution, reproducibility can be accomplished in various ways. For example, if the contribution is a novel architecture, describing the architecture fully might suffice, or if the contribution is a specific model and empirical evaluation, it may be necessary to either make it possible for others to replicate the model with the same dataset, or provide access to the model. In general. releasing code and data is often one good way to accomplish this, but reproducibility can also be provided via detailed instructions for how to replicate the results, access to a hosted model (e.g., in the case of a large language model), releasing of a model checkpoint, or other means that are appropriate to the research performed.
        \item While NeurIPS does not require releasing code, the conference does require all submissions to provide some reasonable avenue for reproducibility, which may depend on the nature of the contribution. For example
        \begin{enumerate}
            \item If the contribution is primarily a new algorithm, the paper should make it clear how to reproduce that algorithm.
            \item If the contribution is primarily a new model architecture, the paper should describe the architecture clearly and fully.
            \item If the contribution is a new model (e.g., a large language model), then there should either be a way to access this model for reproducing the results or a way to reproduce the model (e.g., with an open-source dataset or instructions for how to construct the dataset).
            \item We recognize that reproducibility may be tricky in some cases, in which case authors are welcome to describe the particular way they provide for reproducibility. In the case of closed-source models, it may be that access to the model is limited in some way (e.g., to registered users), but it should be possible for other researchers to have some path to reproducing or verifying the results.
        \end{enumerate}
    \end{itemize}

\item {\bf Open access to data and code}
    \item[] Question: Does the paper provide open access to the data and code, with sufficient instructions to faithfully reproduce the main experimental results, as described in supplemental material?
    \item[] Answer: \answerYes{} % Replace by \answerYes{}, \answerNo{}, or \answerNA{}.
    \item[] Justification:  We provide a benchmarking codebase that makes it easy to reproduce our results at \texttt{github.com/jehanyang/emgbench}, with clear instructions in a \texttt{README.md} document and corresponding configuration files for the results produced in our tables. We also will provide our data for the new dataset in \texttt{emgbench.github.io}. 
    \item[] Guidelines:
    \begin{itemize}
        \item The answer NA means that paper does not include experiments requiring code.
        \item Please see the NeurIPS code and data submission guidelines (\url{https://nips.cc/public/guides/CodeSubmissionPolicy}) for more details.
        \item While we encourage the release of code and data, we understand that this might not be possible, so “No” is an acceptable answer. Papers cannot be rejected simply for not including code, unless this is central to the contribution (e.g., for a new open-source benchmark).
        \item The instructions should contain the exact command and environment needed to run to reproduce the results. See the NeurIPS code and data submission guidelines (\url{https://nips.cc/public/guides/CodeSubmissionPolicy}) for more details.
        \item The authors should provide instructions on data access and preparation, including how to access the raw data, preprocessed data, intermediate data, and generated data, etc.
        \item The authors should provide scripts to reproduce all experimental results for the new proposed method and baselines. If only a subset of experiments are reproducible, they should state which ones are omitted from the script and why.
        \item At submission time, to preserve anonymity, the authors should release anonymized versions (if applicable).
        \item Providing as much information as possible in supplemental material (appended to the paper) is recommended, but including URLs to data and code is permitted.
    \end{itemize}

\item {\bf Experimental Setting/Details}
    \item[] Question: Does the paper specify all the training and test details (e.g., data splits, hyperparameters, how they were chosen, type of optimizer, etc.) necessary to understand the results?
    \item[] Answer: \answerYes{} % Replace by \answerYes{}, \answerNo{}, or \answerNA{}.
    \item[] Justification: Throughout the paper, we provide information on data splits, learning rates, and number of epochs. We further provide a benchmarking codebase that specifies additional information to reproduce our results at \texttt{github.com/jehanyang/emgbench}, with clear instructions in a \texttt{README.md} document and corresponding configuration files for the results produced in our tables. 
    \item[] Guidelines:
    \begin{itemize}
        \item The answer NA means that the paper does not include experiments.
        \item The experimental setting should be presented in the core of the paper to a level of detail that is necessary to appreciate the results and make sense of them.
        \item The full details can be provided either with the code, in appendix, or as supplemental material.
    \end{itemize}

\item {\bf Experiment Statistical Significance}
    \item[] Question: Does the paper report error bars suitably and correctly defined or other appropriate information about the statistical significance of the experiments?
    \item[] Answer: \answerNo{} % Replace by \answerYes{}, \answerNo{}, or \answerNA{}.
    \item[] Justification: Because experiments took approximately 10,000 GPU hours for running each method each time, we do not run multiple tests for each experiment due to limited computational resources. However, we do run Table 3 with 3 seeds. 
    \item[] Guidelines:
    \begin{itemize}
        \item The answer NA means that the paper does not include experiments.
        \item The authors should answer "Yes" if the results are accompanied by error bars, confidence intervals, or statistical significance tests, at least for the experiments that support the main claims of the paper.
        \item The factors of variability that the error bars are capturing should be clearly stated (for example, train/test split, initialization, random drawing of some parameter, or overall run with given experimental conditions).
        \item The method for calculating the error bars should be explained (closed form formula, call to a library function, bootstrap, etc.)
        \item The assumptions made should be given (e.g., Normally distributed errors).
        \item It should be clear whether the error bar is the standard deviation or the standard error of the mean.
        \item It is OK to report 1-sigma error bars, but one should state it. The authors should preferably report a 2-sigma error bar than state that they have a 96\% CI, if the hypothesis of Normality of errors is not verified.
        \item For asymmetric distributions, the authors should be careful not to show in tables or figures symmetric error bars that would yield results that are out of range (e.g. negative error rates).
        \item If error bars are reported in tables or plots, The authors should explain in the text how they were calculated and reference the corresponding figures or tables in the text.
    \end{itemize}

\item {\bf Experiments Compute Resources}
    \item[] Question: For each experiment, does the paper provide sufficient information on the computer resources (type of compute workers, memory, time of execution) needed to reproduce the experiments?
    \item[] Answer: \answerYes{} % Replace by \answerYes{}, \answerNo{}, or \answerNA{}.
    \item[] Justification: We provide this information in Section \ref{section:hardware}. 
    \item[] Guidelines:
    \begin{itemize}
        \item The answer NA means that the paper does not include experiments.
        \item The paper should indicate the type of compute workers CPU or GPU, internal cluster, or cloud provider, including relevant memory and storage.
        \item The paper should provide the amount of compute required for each of the individual experimental runs as well as estimate the total compute. 
        \item The paper should disclose whether the full research project required more compute than the experiments reported in the paper (e.g., preliminary or failed experiments that didn't make it into the paper). 
    \end{itemize}
    
\item {\bf Code Of Ethics}
    \item[] Question: Does the research conducted in the paper conform, in every respect, with the NeurIPS Code of Ethics \url{https://neurips.cc/public/EthicsGuidelines}?
    \item[] Answer: \answerYes{} % Replace by \answerYes{}, \answerNo{}, or \answerNA{}.
    \item[] Justification: All data from participants are anonymized. We do not foresee the benchmarking and use of EMG datasets in our study as violating any part of the NeurIPS Code of Ethics. 
    \item[] Guidelines:
    \begin{itemize}
        \item The answer NA means that the authors have not reviewed the NeurIPS Code of Ethics.
        \item If the authors answer No, they should explain the special circumstances that require a deviation from the Code of Ethics.
        \item The authors should make sure to preserve anonymity (e.g., if there is a special consideration due to laws or regulations in their jurisdiction).
    \end{itemize}

\item {\bf Broader Impacts}
    \item[] Question: Does the paper discuss both potential positive societal impacts and negative societal impacts of the work performed?
    \item[] Answer: \answerYes{} % Replace by \answerYes{}, \answerNo{}, or \answerNA{}.
    \item[] Justification: We mention the positive benefits of improving EMG gesture classification, especially for people with motor impairments in Section \ref{section:emg-as-a-control-interface}. We do not think there is substantial negative societal impacts to mention for benchmarking to create a more robust EMG control interface using learning-based algorithms. 
    \item[] Guidelines:
    \begin{itemize}
        \item The answer NA means that there is no societal impact of the work performed.
        \item If the authors answer NA or No, they should explain why their work has no societal impact or why the paper does not address societal impact.
        \item Examples of negative societal impacts include potential malicious or unintended uses (e.g., disinformation, generating fake profiles, surveillance), fairness considerations (e.g., deployment of technologies that could make decisions that unfairly impact specific groups), privacy considerations, and security considerations.
        \item The conference expects that many papers will be foundational research and not tied to particular applications, let alone deployments. However, if there is a direct path to any negative applications, the authors should point it out. For example, it is legitimate to point out that an improvement in the quality of generative models could be used to generate deepfakes for disinformation. On the other hand, it is not needed to point out that a generic algorithm for optimizing neural networks could enable people to train models that generate Deepfakes faster.
        \item The authors should consider possible harms that could arise when the technology is being used as intended and functioning correctly, harms that could arise when the technology is being used as intended but gives incorrect results, and harms following from (intentional or unintentional) misuse of the technology.
        \item If there are negative societal impacts, the authors could also discuss possible mitigation strategies (e.g., gated release of models, providing defenses in addition to attacks, mechanisms for monitoring misuse, mechanisms to monitor how a system learns from feedback over time, improving the efficiency and accessibility of ML).
    \end{itemize}
    
\item {\bf Safeguards}
    \item[] Question: Does the paper describe safeguards that have been put in place for responsible release of data or models that have a high risk for misuse (e.g., pretrained language models, image generators, or scraped datasets)?
    \item[] Answer: \answerNA{} % Replace by \answerYes{}, \answerNo{}, or \answerNA{}.
    \item[] Justification: We expect that EMG benchmarking and the datasets have a low risk for misuse. 
    \item[] Guidelines:
    \begin{itemize}
        \item The answer NA means that the paper poses no such risks.
        \item Released models that have a high risk for misuse or dual-use should be released with necessary safeguards to allow for controlled use of the model, for example by requiring that users adhere to usage guidelines or restrictions to access the model or implementing safety filters. 
        \item Datasets that have been scraped from the Internet could pose safety risks. The authors should describe how they avoided releasing unsafe images.
        \item We recognize that providing effective safeguards is challenging, and many papers do not require this, but we encourage authors to take this into account and make a best faith effort.
    \end{itemize}

\item {\bf Licenses for existing assets}
    \item[] Question: Are the creators or original owners of assets (e.g., code, data, models), used in the paper, properly credited and are the license and terms of use explicitly mentioned and properly respected?
    \item[] Answer: \answerYes{} % Replace by \answerYes{}, \answerNo{}, or \answerNA{}.
    \item[] Justification: All data is used under their existing licenses. All code released is created by the authors. Data is linked to original dataset sources as citations or as urls in the codebase. 
    \item[] Guidelines:
    \begin{itemize}
        \item The answer NA means that the paper does not use existing assets.
        \item The authors should cite the original paper that produced the code package or dataset.
        \item The authors should state which version of the asset is used and, if possible, include a URL.
        \item The name of the license (e.g., CC-BY 4.0) should be included for each asset.
        \item For scraped data from a particular source (e.g., website), the copyright and terms of service of that source should be provided.
        \item If assets are released, the license, copyright information, and terms of use in the package should be provided. For popular datasets, \url{paperswithcode.com/datasets} has curated licenses for some datasets. Their licensing guide can help determine the license of a dataset.
        \item For existing datasets that are re-packaged, both the original license and the license of the derived asset (if it has changed) should be provided.
        \item If this information is not available online, the authors are encouraged to reach out to the asset's creators.
    \end{itemize}

\item {\bf New Assets}
    \item[] Question: Are new assets introduced in the paper well documented and is the documentation provided alongside the assets?
    \item[] Answer: \answerYes{} % Replace by \answerYes{}, \answerNo{}, or \answerNA{}.
    \item[] Justification: Details on the methods for FlexWear-HD dataset collection for the new dataset are mentioned in \cite{yang2023high} and Appendix \ref{appendix:flexwear-hd}. 
    \item[] Guidelines:
    \begin{itemize}
        \item The answer NA means that the paper does not release new assets.
        \item Researchers should communicate the details of the dataset/code/model as part of their submissions via structured templates. This includes details about training, license, limitations, etc. 
        \item The paper should discuss whether and how consent was obtained from people whose asset is used.
        \item At submission time, remember to anonymize your assets (if applicable). You can either create an anonymized URL or include an anonymized zip file.
    \end{itemize}

\item {\bf Crowdsourcing and Research with Human Subjects}
    \item[] Question: For crowdsourcing experiments and research with human subjects, does the paper include the full text of instructions given to participants and screenshots, if applicable, as well as details about compensation (if any)? 
    \item[] Answer: \answerYes{} % Replace by \answerYes{}, \answerNo{}, or \answerNA{}.
    \item[] Justification: Details on the methods for FlexWear-HD dataset collection for the new dataset are mentioned in \cite{yang2023high} and Appendix \ref{appendix:flexwear-hd}. Workers involved in data collection are paid at least the minimum wage in the United States. 
    \item[] Guidelines:
    \begin{itemize}
        \item The answer NA means that the paper does not involve crowdsourcing nor research with human subjects.
        \item Including this information in the supplemental material is fine, but if the main contribution of the paper involves human subjects, then as much detail as possible should be included in the main paper. 
        \item According to the NeurIPS Code of Ethics, workers involved in data collection, curation, or other labor should be paid at least the minimum wage in the country of the data collector. 
    \end{itemize}

\item {\bf Institutional Review Board (IRB) Approvals or Equivalent for Research with Human Subjects}
    \item[] Question: Does the paper describe potential risks incurred by study participants, whether such risks were disclosed to the subjects, and whether Institutional Review Board (IRB) approvals (or an equivalent approval/review based on the requirements of your country or institution) were obtained?
    \item[] Answer: \answerYes{} % Replace by \answerYes{}, \answerNo{}, or \answerNA{}.
    \item[] Justification: We link to the IRB approval statement in Appendix \ref{appendix:flexwear-hd}. 
    \item[] Guidelines:
    \begin{itemize}
        \item The answer NA means that the paper does not involve crowdsourcing nor research with human subjects.
        \item Depending on the country in which research is conducted, IRB approval (or equivalent) may be required for any human subjects research. If you obtained IRB approval, you should clearly state this in the paper. 
        \item We recognize that the procedures for this may vary significantly between institutions and locations, and we expect authors to adhere to the NeurIPS Code of Ethics and the guidelines for their institution. 
        \item For initial submissions, do not include any information that would break anonymity (if applicable), such as the institution conducting the review.
    \end{itemize}

\end{enumerate}

\end{document}